\pdfoutput=1
\documentclass[11pt]{article}
\usepackage[T1]{fontenc}
\usepackage[utf8]{inputenc}
\usepackage{lmodern}
\usepackage{mathpazo}
\usepackage[a4paper,margin=2.2cm,top=2.6cm,bottom=2.4cm,headheight=14pt]{geometry}
\usepackage{amsmath,amssymb}
\usepackage{graphicx}
\usepackage{booktabs,tabularx,array,longtable,calc}
\usepackage{caption}
\DeclareCaptionLabelSeparator{pipe}{\ \textbar\ }
\usepackage{titlesec}
\titleformat{\section}{\large\bfseries}{\thesection.}{0.5em}{}
\titleformat{\subsection}{\normalsize\bfseries}{\thesubsection.}{0.5em}{}
\titleformat{\subsubsection}{\normalsize\bfseries}{\thesubsubsection.}{0.5em}{}
\titlespacing*{\section}{0pt}{1.8em}{0.8em}
\titlespacing*{\subsection}{0pt}{1.4em}{0.6em}
\titlespacing*{\subsubsection}{0pt}{1.2em}{0.5em}
\usepackage{fancyhdr}
\fancypagestyle{plain}{\fancyhf{}\fancyfoot[C]{\thepage}}
\usepackage{microtype}
\usepackage[hidelinks]{hyperref}
\DeclareUnicodeCharacter{00B0}{\ensuremath{^\circ}}
\DeclareUnicodeCharacter{0131}{\i{}}
\DeclareUnicodeCharacter{0394}{\ensuremath{\Delta}}
\DeclareUnicodeCharacter{03B4}{\ensuremath{\delta}}
\DeclareUnicodeCharacter{03B5}{\ensuremath{\varepsilon}}
\DeclareUnicodeCharacter{2020}{\textdagger{}}
\DeclareUnicodeCharacter{2113}{\ensuremath{\ell}}
\DeclareUnicodeCharacter{2212}{\ensuremath{-}}
\DeclareUnicodeCharacter{2264}{\ensuremath{\le}}
\DeclareUnicodeCharacter{2265}{\ensuremath{\ge}}

\providecommand{\tightlist}{\setlength{\itemsep}{0pt}\setlength{\parskip}{0pt}}
\renewenvironment{abstract}{\begin{center}{\Large\bfseries Abstract}\end{center}\vspace{-0.2em}%
  \begin{list}{}{\setlength{\leftmargin}{0.14\textwidth}\setlength{\rightmargin}{0.14\textwidth}}\item\relax}%
  {\end{list}\vspace{1em}}
\hypersetup{pdftitle={HistGramBench: Measuring Structural Recovery When Multimodal Models Redraw Historical Mathematical Diagrams as TikZ},pdfauthor={Minh Quang Tran; Carter He; Cameron Bloch; Zitai Huang; Thien Truong Vo; Trong Nguyen Tran; Isaac Chen}}
\begin{document}
\thispagestyle{plain}
\begin{center}
\vspace*{0.6em}
{\LARGE\bfseries HistGramBench: Measuring Structural Recovery When Multimodal Models Redraw Historical Mathematical Diagrams as TikZ\par}
\vspace{1.4em}
{\bfseries \mbox{Minh Quang Tran$^{1,*}$}\quad \mbox{Carter He$^{1}$}\quad \mbox{Cameron Bloch$^{1}$}\quad \mbox{Zitai Huang$^{1}$}\quad \mbox{Thien Truong Vo$^{2}$}\quad \mbox{Trong Nguyen Tran$^{2}$}\quad \mbox{Isaac Chen$^{1}$}\par}\vspace{0.3em}{\small\bfseries $^{1}$Deerfield Academy, Deerfield, Massachusetts, USA\par}{\small\bfseries $^{2}$University of Economics Ho Chi Minh City (UEH), Ho Chi Minh City, Vietnam\par}\vspace{0.5em}{\small $^{*}$Corresponding author: \href{mailto:qtran28@deerfield.edu}{\texttt{qtran28@deerfield.edu}}\par}\vspace{0.2em}{\small ORCID: \mbox{\href{https://orcid.org/0009-0004-7852-0193}{\texttt{0009-0004-7852-0193}} (Minh Quang Tran)}, \mbox{\href{https://orcid.org/0009-0006-7906-8164}{\texttt{0009-0006-7906-8164}} (Carter He)}, \mbox{\href{https://orcid.org/0009-0008-3829-9964}{\texttt{0009-0008-3829-9964}} (Cameron Bloch)}, \mbox{\href{https://orcid.org/0009-0004-8832-2350}{\texttt{0009-0004-8832-2350}} (Zitai Huang)}, \mbox{\href{https://orcid.org/0009-0008-9633-5775}{\texttt{0009-0008-9633-5775}} (Thien Truong Vo)}, \mbox{\href{https://orcid.org/0009-0006-5782-4918}{\texttt{0009-0006-5782-4918}} (Trong Nguyen Tran)}, \mbox{\href{https://orcid.org/0009-0006-8984-5772}{\texttt{0009-0006-8984-5772}} (Isaac Chen)}\par}
\end{center}
\vspace{0.6em}

\begin{abstract}

Digital libraries preserve mathematical books printed from the sixteenth to the twentieth century, but the lettered diagrams in them, the figures on which Euclidean proofs depend, remain scanned images whose content cannot be searched. Redrawing such a diagram as a short drawing program would record which points lie on which lines and circles. But a printed plate offers no source code against which the redrawing could be checked. We introduce HistGramBench, a benchmark of 2,507 diagrams from 95 works printed between 1551 and 1933 in eight languages. For each diagram, two AI models from different providers agree on the facts a reader takes from the figure, such as which lines are parallel or where two lines meet, and every redrawing is checked against those facts on the model's own drawing. We evaluate six affordable AI models. The best recovers about three fifths of these facts, and most of what the models miss comes from programs that never produce a drawing. Parallels and tangencies are the hardest relations to reproduce. HistGramBench gives historians and librarians a way to measure when such tools can make old mathematical diagrams searchable.

\textbf{Keywords:} history of mathematics; historical mathematical diagrams; vision-language models; TikZ; graphics program synthesis; structural evaluation; digital mathematics libraries

\end{abstract}

\section{Introduction}\label{introduction}

A page from a nineteenth-century geometry textbook contains two kinds of content. Its text can be transcribed, searched and translated with mature tools. The figure beside it, a circle with a few lettered points and the lines that join them, remains a picture. Yet that picture carries the mathematical relations the text requires: which point lies on the circle, which lines meet, and which letter designates each object. Digital libraries have preserved these figures as images. In this paper we measure how far current multimodal models can convert them into structure that a machine can read and check.

\subsection{Diagrams in Printed Mathematics and Their Digitization}\label{diagrams-in-printed-mathematics-and-their-digitization}

In printed mathematics the lettered diagram is part of the argument. Greek deductive practice was built around it; the letters of the text name points in the figure, and the proof moves between the two (Netz, 1999). Euclidean proofs also read some properties directly off the figure (Manders, 2008). From the drawing, the reader takes incidence (a point lies on a circle), betweenness (a point lies between two others on a line) and containment (one region lies inside another).

Most figures studied here come from the late nineteenth and early twentieth centuries, and drawing conventions vary across the corpus (Figure 1). Most works letter their points and lines, whereas Byrne's 1847 edition of Euclid identifies objects by colour (Byrne, 1847), a convention outside the label vocabulary defined in Section 5.1.

Digital mathematics libraries such as Numdam and the Göttingen digitization center store these works as page images. The proposal for a global digital mathematics library asks for a corpus that machines can act on as well as display (Bouche, 2008; Committee on Planning a Global Library of the Mathematical Sciences, 2014). Text and formulas already have recognition pipelines (Blecher et al., 2024; Suzuki et al., 2003), and work on historical diagrams recovers geometric primitives and text regions (Baltacı et al., 2026; Kalleli et al., 2024), but none of these pipelines returns a figure's relations and labels as an editable, queryable object. Redrawing the figure as code is one route to such an object.

\begin{figure}[!htbp]
\centering
\includegraphics[width=\linewidth]{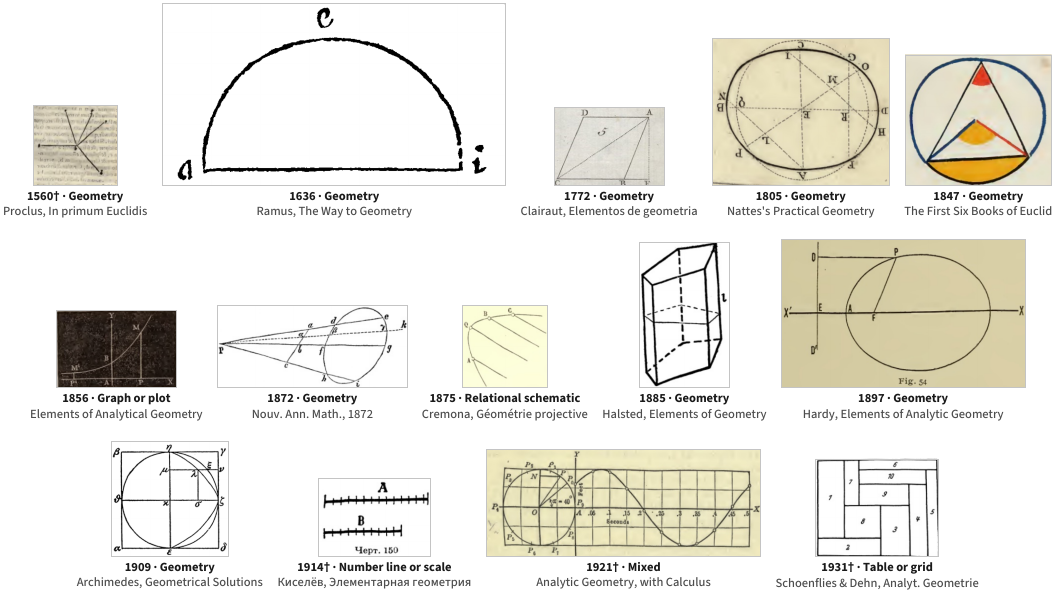}
\caption{Sample of corpus figures, 1560--1931, with year, figure type and source. †: not in the scored test set. Each crop is shown as reproduced by the repository it was taken from, which for Project Gutenberg is a transcriber-prepared image of the printed plate.}
\label{fig:1}
\end{figure}

\subsection{Graphics Program Synthesis and the Evaluation Problem}\label{graphics-program-synthesis-and-the-evaluation-problem}

TikZ (Tantau, 2023) is a widely used LaTeX drawing language in which a figure is a short program of named points, paths and labels (Belouadi et al., 2025; Belouadi, Lauscher, \& Eger, 2024). A figure redrawn in TikZ expresses its structure explicitly, as the program itself states that the line through A and B meets the circle at C. Vision-language models already write such programs from images (Belouadi, Ponzetto, \& Eger, 2024). What remains open is how to judge the redrawing of a figure that has no source code.

Three obvious checks fall short. A compiling program shows only that it runs and says nothing about whether it draws the figure. Pixel similarity between the render and the scan penalizes harmless differences in size, position and stroke, which has led image-to-vector studies to reject pixel metrics (Rodriguez et al., 2025). Comparison with reference code is impossible, because a printed plate has none.

The alternative is to check, on the model's own drawing, the facts that a reader of the figure uses. If the key says that point C lies on the circle, the test asks whether the drawn point labelled C lies on the drawn circle, wherever the model has placed both. Caption evaluation gained agreement with human judgment in the same way, by scoring the propositions a caption asserts instead of its n-grams (Anderson et al., 2016). A check of this kind is indifferent to scale, position and style, and it needs no reference program.

\subsection{Contributions, Research Questions and Findings}\label{contributions-research-questions-and-findings}

We make four contributions.

\begin{enumerate}
\def\labelenumi{\arabic{enumi}.}
\tightlist
\item
  \textbf{Corpus.} 2,507 scored figures from 95 public-domain mathematical works printed between 1551 and 1933, screened, deduplicated and split by source group so that no test figure shares a work or a near-identical plate with the development data.
\item
  \textbf{Answer key.} For each figure, a set of typed facts in three relation tiers, admitted only when two model writers from different providers both support it.
\item
  \textbf{Scoring.} A protocol that compiles each answer once, reads the drawing as vector geometry, checks every fact whose objects bind with tolerances scaled to the drawing, and sends the rest to judge models from other families.
\item
  \textbf{Evaluation.} Six low-cost vision-language models, with and without page text, analysed at the level of source groups with false-discovery control, a pre-registered equivalence rule and ten scoring and weighting variants.
\end{enumerate}

The evaluation answers four research questions (RQ1--RQ4). RQ1 asks how much of the answer key the six models recover as compilable TikZ, and how much of the loss comes from answers with no drawing. RQ2 asks where the losses fall, by class and tier of structure and by kind of figure. RQ3 asks whether adding the page text around the figure changes recovery, pooled over models and for individual models, and RQ4 tests whether the order and rank bands of the models remain stable under changes to the judges, the scoring rules, the weighting and the multiplicity correction.

On the image alone, the six models recover between 0.20 and 0.59 of the answer key. Between 27 and 49 percent of their answers yield no drawing, and these missing drawings account for 58 to 85 percent of each model's loss. Relations are the hardest facts for every model, with depicted exact relations (parallel, perpendicular, tangent, collinear, concyclic) harder than co-exact ones. Pooled over the six models, page text leaves recovery unchanged within the pre-registered margin. The point order of the six models is identical across ten scoring and weighting variants, up to one pair that the headline already ties, and every variant that tests separation separates the same three tiers: \{Gemma 4 26B, Llama 4 Maverick\}, then \{Ministral 3 14B, Gemma 4 E2B, Claude Haiku 4.5\}, then \{Nemotron Nano 12B v2\}.

Section 2 reviews prior work, and Sections 3 to 7 define the corpus, task, answer key, scoring and setup. Section 8 reports the results, Section 9 discusses them and Section 10 concludes.

\section{Related Work}\label{related-work}

\subsection{Exact and Co-exact Properties in Diagrammatic Reasoning}\label{exact-and-co-exact-properties-in-diagrammatic-reasoning}

The philosophy and history of mathematical practice separate the properties of a diagram that an argument may read off the figure from those it must assert in the text, and this distinction structures the fact vocabulary of HistGramBench. Netz (1999) shows that the lettered diagram is constitutive of Greek deduction, in that the diagram fixes the objects about which the text then reasons. Manders (2008) names the two kinds of property. Co-exact properties, among them incidence, intersection, containment, betweenness and the arrangement of regions, remain stable under small drawing errors and can be read directly from the diagram. Exact properties, such as equal lengths and angles, parallelism, tangency, straightness and concyclicity, are guaranteed by no drawing and are asserted in the text. Hamami et al.~(2020) list parallelism, tangency and concyclicity among the exact relations and test this account experimentally with human readers.

Avigad et al.~(2009) and Mumma (2010) show that a finite vocabulary of co-exact relations formally captures the diagrammatic inferences in Euclid's \emph{Elements}. Saito and Sidoli (2012) find that manuscript diagrams differ from the redrawn figures of modern critical editions, which regularize them. Redrawing is therefore an interpretive act that can erase information. Nineteenth-century axiomatics moved the authority for order and incidence from the figure to the axiom (Hilbert, 1899; Pasch, 1882). We nonetheless treat the printed textbook figures that dominate this corpus as scale drawings, whose depicted parallels and tangencies are part of what the reader sees.

The vocabulary used here keeps three tiers apart. Co-exact relations are checked as Manders describes them. Depicted exact relations are checked under an explicit tolerance, because a TikZ drawing realizes them exactly and a scale drawing shows them. Exact equalities count only when the figure marks them, for example with tick marks. The position that a faithful redrawing preserves all three tiers of the printed figure is our own. The theory applies to synthetic-geometry figures, and the 404 graphs and plots in the corpus are scored on the parts of the vocabulary that apply to them, such as axes, curves, incidence and labels.

\subsection{Image-to-Code and Graphics Program Benchmarks}\label{image-to-code-and-graphics-program-benchmarks}

Image-to-code benchmarks score appearance on modern material with known code. In TikZ synthesis, AutomaTikZ (Belouadi, Lauscher, \& Eger, 2024) and TikZero (Belouadi et al., 2025) generate TikZ from text. DeTikZify (Belouadi, Ponzetto, \& Eger, 2024), the image-conditioned member of the family, is evaluated by image and code similarity and by human ratings. For plots and charts, Plot2Code (Wu et al., 2025) and ChartMimic (Yang et al., 2025) report pass rates, code match and model-rated similarity of renders; for web pages, Design2Code (Si et al., 2025) matches visual blocks between screenshot and render. In vector graphics, StarVector (Rodriguez et al., 2025) argues that pixel metrics fail, and VGBench (Zou et al., 2024) finds that models handle high-level formats such as TikZ better than Scalable Vector Graphics (SVG).

Several benchmarks from 2025 and 2026 come closer to the present design. Vision2Code (Periasami et al., 2026) needs no reference code, includes geometry, and separates render success from reconstruction quality with a model rater and deterministic guardrails. ``Solving Is Not Drawing'' (Peng et al., 2026) and GeoBuildBench (Kim et al., 2026) check drawn geometry against specified constraints, but their figures are specified in text without reference to source images. VFIG (He et al., 2026) evaluates figure vectorization with structure-aware measures, and DiagramIR (Kumar et al., 2025) scores generated geometry through an intermediate representation that a language model derives from the code. At the level of relations, ScImage (L. Zhang et al., 2025) judges text-to-figure output with human raters, and MagicGeo (T. Zhang et al., 2026) solves for coordinates that satisfy stated constraints. Concurrent work finds diagram-to-code harder than question answering on the same modern diagrams (Bo et al., 2026).

No benchmark among those reviewed here combines all three conditions: a historical source image, an absence of reference code, and deterministic relation and label checks on the compiled geometry of the model's own drawing. Of the work reviewed above, the image-to-code benchmarks meet at most the second and third conditions, and the work on historical diagrams (Baltacı et al., 2026; Kalleli et al., 2024) meets the first but does not score reconstructions. Vision2Code is the nearest prior work, and Section 8.2 credits its separation of render success from reconstruction quality.

\subsection{Geometry Diagram Parsing and Visual Mathematical Reasoning}\label{geometry-diagram-parsing-and-visual-mathematical-reasoning}

Geometry-diagram parsing supplies the typed-relation representation. Its benchmarks use clean modern diagrams with human annotation and score answers or parses without requiring code reconstruction. GeoS couples diagram and text interpretation to solve geometry questions (Seo et al., 2015). Inter-GPS annotates Geometry3K as formal literals (Lu et al., 2021), and PGDP5K provides primitive and relation annotations (M.-L. Zhang et al., 2022).

The HistGramBench vocabulary maps onto these annotations, as Appendix B details. Incidence, parallel and perpendicular literals map one to one, and PGDP5K's text-to-primitive relation is the parsing precedent for the label facts. The numeric length and angle literals of Inter-GPS are excluded by design, because coordinates and measures are never facts here (Section 5.1.2). FormalGeo (X. Zhang et al., 2023) includes constructions that this vocabulary leaves to the text, such as midpoints and angle bisectors stated without a mark in the figure.

Benchmarks of visual mathematics evaluate answers about diagrams. MathVista (Lu et al., 2024) collects such questions across sources, and MathVerse (R. Zhang et al., 2024) varies how much of the needed information appears in the text versus the image, finding that models often answer without reading the diagram. HistGramBench keeps the typed-relation representation of diagram parsing and moves it from parsing and question answering to the reconstruction of printed historical figures.

\subsection{Historical Document Analysis and Model-Based Evaluation}\label{historical-document-analysis-and-model-based-evaluation}

Document analysis recovers layout, text and, for historical diagrams, primitives and text regions. Graphics recognition and vectorization recover strokes and arcs from line drawings (Egiazarian et al., 2020; Hilaire \& Tombre, 2006); Kalleli et al.~(2024) detect geometric primitives in historical astronomical diagrams, and Baltacı et al.~(2026) detect their text regions. Layout benchmarks and parsers, such as DocLayNet (Pfitzmann et al., 2022), OmniDocBench (Ouyang et al., 2025) and PaddleOCR-VL (Cui et al., 2025), place figure boxes on pages. Datasets of historical documents centre on classification, layout and text (Nikolaidou et al., 2022), and model evaluation on historical print measures transcription (Levchenko, 2025). No historical work we know of recovers typed relations and label attachment, or scores a model's redrawing.

Model judges make large evaluations possible but introduce known biases. Zheng et al.~(2023) document position, verbosity and self-enhancement biases, and Panickssery et al.~(2024) show that a model recognizes and prefers its own generations. Verga et al.~(2024) find that a panel of smaller judges from different families agrees with human judgment better than one large judge. We therefore limit what judges decide, keep their families apart from those of the writers and the contestant, and report how much the results depend on them. Because items cluster within works, inference uses the source group as the unit (Cameron \& Miller, 2015; Miller, 2024).

\section{The HistGramBench Corpus}\label{the-histgrambench-corpus}

\subsection{Source Repositories and Rights}\label{source-repositories-and-rights}

We draw candidate works from Wikimedia Commons, the Internet Archive, the Göttingen digitization center (GDZ), Numdam and Project Gutenberg, together with two archival scans (Appendix D). A work enters the corpus when its evidenced first publication falls in or before 1930 and it was published rather than circulated in manuscript. All 95 works that contribute scored figures are in the public domain; three of them, printed between 1931 and 1933, entered on a recorded rights determination (Appendix D).

\subsection{Figure Extraction and Screening}\label{figure-extraction-and-screening}

We cut figures from pages in three steps (Figure 2, left lane). First, a layout model, PaddleOCR-VL (Cui et al., 2025), proposes figure regions on page scans; where a repository already supplies plate images, the plate itself is taken as the item and no region is proposed.

Second, image gates remove crops that cannot be scored fairly, such as regions that are too small, photographic or mostly text, and split multi-panel regions into their panels. Every removal records one reason from a closed vocabulary, and Table S1 counts the exclusions of every stage by reason.

Third, a model screener admits a crop only when it shows a legible mathematical figure that TikZ can express, rejecting it when in doubt. Screening precedes every contestant answer, because a figure that cannot be read or drawn in TikZ would measure the screener's luck instead of the contestant's ability. We then cap the number of figures that any one group contributes, so that a single long book cannot dominate the test set.

\begin{figure}[!htbp]
\centering
\includegraphics[width=\linewidth]{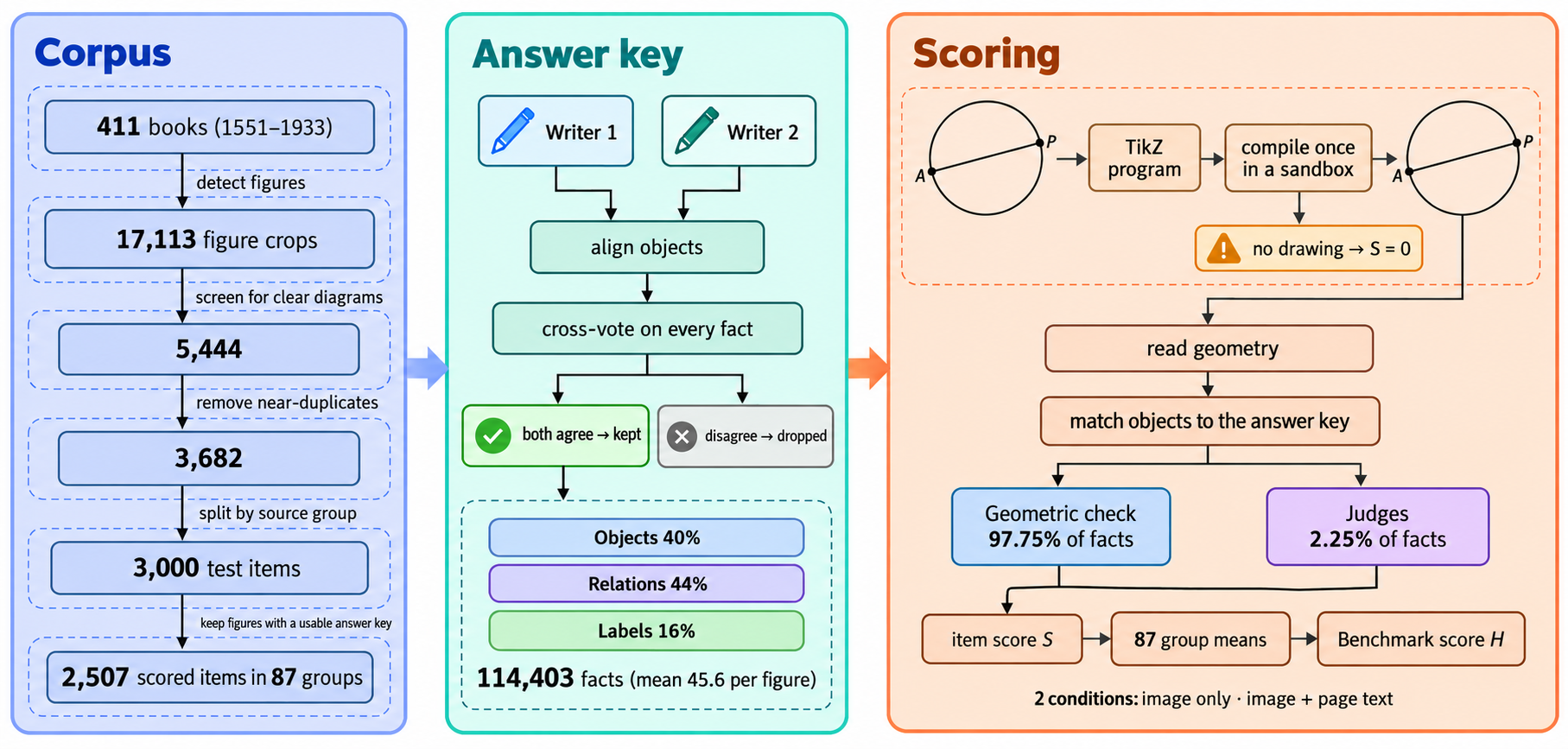}
\caption{Benchmark pipeline. Left: from page scans to scored figures, with counts at each stage. Centre: answer-key construction from two writers of different providers under the unanimity rule. Right: scoring of one compiled answer, in which each key fact is decided on the compiled geometry when its objects bind and by the judge panel otherwise; judges decide about a third of the verdicts on valid drawings.}
\label{fig:2}
\end{figure}

\subsection{Deduplication, Source Groups and Data Split}\label{deduplication-source-groups-and-data-split}

We cluster near-identical crops, such as reprints, second scans of the same plate and repeated figures within a book, and represent each cluster once, counting two crops as near-identical when their 64-bit perceptual hashes lie within Hamming distance 6.

Works are then linked into source groups. Two works belong to the same group when they are the same work, share an author and title, or contain a near-duplicate crop of each other, and each group falls entirely on one side of the development and test split, so near-identical plates never appear on both sides. The development side contains 150 figures from five works and fixed every scoring level, tolerance and threshold before the test run. The test side comprises 2,507 scored figures in 87 groups, the largest of which has 669 figures; Section 8.5 shows that removing it moves no score by more than 0.002.

The source group is also the unit of inference (Section 6.4), because figures from one work share a printer, an engraver and a set of drawing conventions, and treating them as independent observations would overstate the precision of every comparison.

\subsection{Corpus Composition}\label{corpus-composition}

Table 1 gives the composition of the scored set, and Figure S1 shows period by repository. Of the 56 journal articles and extracts, 53 come from the \emph{Nouvelles annales de mathématiques}, held by Numdam. The figures reach the benchmark through three digitization pathways: crops cut by the layout model from page scans, plate images supplied by repositories, and transcriber-prepared images from Project Gutenberg. Ink touches the border of the crop on 824 figures. Page text is present for 1,538 figures and absent for 969 plates, most of them repository plates supplied without their page; Section 8.4 compares the two sets.

The headline score weights every source group equally (Section 6.4). Under this weighting the 56 single-work journal groups represent 64 percent of the headline weight while holding 142 figures, and the 30 groups that contain a single figure account for 34.5 percent. Section 8.5 therefore reports the headline without the Numdam groups and with each figure weighted equally.

\begin{table}[t]
\centering
\caption{Composition of the scored set ($N =$ 2,507 figures, 95 works, $G = 87$ source groups). Weight: share of the headline $H$, which weights every source group equally.}
\label{tab:composition}
\small
\begin{tabular}{@{}l rrrr@{}}
\toprule
 & Figures & Works & Groups & Weight (\%) \\
\midrule
\multicolumn{5}{@{}l}{\textit{Period}} \\
\quad Before 1850 & 375 & 7 & 7 & 5.8 \\
\quad 1850–1930 & 1,965 & 85 & 82 & 92.5 \\
\quad After 1930 & 167 & 3 & 3 & 1.7 \\
\addlinespace
\multicolumn{5}{@{}l}{\textit{Genre}} \\
\quad Book & 2,365 & 39 & 31 & 35.6 \\
\quad Journal article & 142 & 56 & 56 & 64.4 \\
\addlinespace
\multicolumn{5}{@{}l}{\textit{Language of the work}} \\
\quad English & 1,474 & 23 & 19 & 20.9 \\
\quad German & 315 & 10 & 9 & 10.3 \\
\quad Russian & 204 & 2 & 2 & 1.4 \\
\quad French & 183 & 56 & 56 & 64.4 \\
\quad Chinese & 133 & 1 & 1 & 0.2 \\
\quad Italian & 115 & 1 & 1 & 1.1 \\
\quad Portuguese & 44 & 1 & 1 & 1.1 \\
\quad Gujarati & 39 & 1 & 1 & 0.4 \\
\addlinespace
\multicolumn{5}{@{}l}{\textit{Digitization pathway}} \\
\quad Page-scan crop & 1,568 & 26 & 22 & 24.6 \\
\quad Repository plate image & 678 & 64 & 62 & 70.6 \\
\quad Transcriber-prepared image & 261 & 5 & 5 & 4.8 \\
\addlinespace
\multicolumn{5}{@{}l}{\textit{Repository}} \\
\quad Internet Archive & 948 & 18 & 16 & 17.9 \\
\quad Wikimedia Commons & 620 & 8 & 7 & 6.8 \\
\quad GDZ Göttingen & 294 & 9 & 8 & 9.2 \\
\quad Archival & 265 & 2 & 1 & 0.5 \\
\quad Project Gutenberg & 261 & 5 & 5 & 4.8 \\
\quad Numdam & 119 & 53 & 53 & 60.9 \\
\addlinespace
\multicolumn{5}{@{}l}{\textit{Figure type}} \\
\quad Geometry & 2,079 & 92 & 84 & 84.4 \\
\quad Graphs and plots & 404 & 39 & 34 & 15.0 \\
\quad Other & 24 & 13 & 11 & 0.5 \\
\midrule
Total & 2,507 & 95 & 87 & 100.0 \\
\bottomrule
\end{tabular}
\end{table}

\section{Task Definition}\label{task-definition}

\subsection{Evaluation Conditions}\label{evaluation-conditions}

We pose one task under two conditions (Table 2). In the image-only condition, Track A, the model receives the figure and the instruction to reconstruct ``the mathematical figure the original author intended'', without reproducing scan damage, paper texture or print defects. The target is therefore the printed witness: the figure as it stands in the digitized copy, with reproduction damage removed and its content kept as printed, uncorrected against the text. The answer key accordingly records the printed result of the combined work of author, draughtsman, engraver and editor.

In the image-plus-text condition, Track B, the same prompt is followed by up to 1,200 characters of text from the same page, caption first. Item, answer key, judges and scoring are shared. The paired difference between the tracks measures what the page text contributes. For the 969 plates without page text, the Track B prompt contains an empty context header. These plates serve as a control, since on them the input differs only by the wrapper and any difference between the tracks comes from the wrapper or from run-to-run variation of the model.

\begin{table}[!htbp]
\centering
\caption{The two task conditions; Appendix A gives the full prompts.}\label{tab:2}
{\small\setlength{\tabcolsep}{3pt}
\begin{tabular}{@{}
  >{\raggedright\arraybackslash}p{(\linewidth - 4\tabcolsep) * \real{0.3333}}
  >{\raggedright\arraybackslash}p{(\linewidth - 4\tabcolsep) * \real{0.3333}}
  >{\raggedright\arraybackslash}p{(\linewidth - 4\tabcolsep) * \real{0.3333}}@{}}
\toprule\noalign{}
\begin{minipage}[b]{\linewidth}\raggedright
\end{minipage} & \begin{minipage}[b]{\linewidth}\raggedright
Track A (image only)
\end{minipage} & \begin{minipage}[b]{\linewidth}\raggedright
Track B (image plus page text)
\end{minipage} \\
\midrule\noalign{}
Input & Figure crop and instruction & Same, plus up to 1,200 characters of page text, caption first \\
Output & One TikZ picture & Same \\
Answer key, judges, scoring & Shared & Shared \\
Figures with page text & --- & 1,538 of 2,507 \\
\bottomrule
\end{tabular}}
\end{table}

\subsection{Output Specification and Invariances}\label{output-specification-and-invariances}

An answer must contain exactly one fenced LaTeX block holding exactly one \texttt{tikzpicture}, optionally inside a standalone document. Packages and TikZ libraries come from a published allowlist, and commands that read or write files, run shell code, include images or select fonts are forbidden (Appendix A). Labels may be written in their original script, since the harness supplies fonts for every script.

The prompt also states explicitly which dimensions are excluded from evaluation: translation, uniform scale, line width, font, colour, label offset within tolerance, canvas padding, and any rewriting of the code that renders the same figure. For facts decided on the geometry, each stated invariance is a property of the test itself, since every test compares distances and angles relative to the size of the drawing (Section 6.3.1); for facts decided by judges, the panel receives the same list as an instruction.

\section{Answer-Key Construction}\label{answer-key-construction}

\subsection{Fact Representation}\label{fact-representation}

\subsubsection{Fact Classes and Relation Tiers}\label{fact-classes-and-relation-tiers}

Each figure's answer key is a set of typed facts in three classes, and Table 3 lists every predicate. Entity facts state that an object exists, with 16 entity types. Relation facts state how objects stand to one another, with 20 types divided into three tiers and a residual group: the co-exact tier; the depicted exact tier, checked under tolerance; the marked exact tier, whose equalities count only when the figure marks them; and three-dimensional and other relations, which judges decide. Label facts state which printed string names which object (label) and which printed measures appear (measure\_label).

Three conventions make the facts well defined. A point exists only where the figure labels it or marks it with a dot, so the unlabelled crossing of two lines is not a point entity, though it can be the argument of intersect\_at. A straight stroke is typed as a segment, ray or line according to where its labelled points lie along it. A polygon is a chordless cycle of at most eight corners, so a parallelogram drawn with a diagonal yields its two triangles and no parallelogram.

\begin{table}[!htbp]
\centering
\caption{Fact vocabulary of the answer key (n = 114,403 key facts). Geometry: decided on the compiled drawing when every argument binds, otherwise by the judges. Share: percent of all key facts. Appendix B gives the arguments and geometric test for every predicate.}\label{tab:3}
{\small\setlength{\tabcolsep}{3pt}
\begin{tabular}{@{}
  >{\raggedright\arraybackslash}p{(\linewidth - 6\tabcolsep) * \real{0.2500}}
  >{\raggedright\arraybackslash}p{(\linewidth - 6\tabcolsep) * \real{0.2500}}
  >{\raggedright\arraybackslash}p{(\linewidth - 6\tabcolsep) * \real{0.2500}}
  >{\raggedright\arraybackslash}p{(\linewidth - 6\tabcolsep) * \real{0.2500}}@{}}
\toprule\noalign{}
\begin{minipage}[b]{\linewidth}\raggedright
Class and tier
\end{minipage} & \begin{minipage}[b]{\linewidth}\raggedright
Predicates
\end{minipage} & \begin{minipage}[b]{\linewidth}\raggedright
Decided by
\end{minipage} & \begin{minipage}[b]{\linewidth}\raggedright
Share (\%)
\end{minipage} \\
\midrule\noalign{}
Entity & point, segment, line, ray, circle, arc, ellipse, polygon, axis, arrow, angle marker, right-angle mark, length tick & Geometry & 38.5 \\
Entity & region, curve, solid & Judges & 1.3 \\
Relation, co-exact & on, endpoint\_of, vertex\_of, center\_of, edge\_of, arc\_of, intersect\_at, between, contains & Geometry & 39.0 \\
Relation, depicted exact & parallel, perpendicular, tangent\_to, collinear, concyclic & Geometry & 5.0 \\
Relation, marked exact & equal\_length, equal\_angle & Judges, when marked & 0.06 \\
Relation, three-dimensional and other & face\_of, cross\_section\_of, hidden\_edge, asymptote\_of & Judges & 0.3 \\
Label & label, measure\_label & Geometry & 15.7 \\
\bottomrule
\end{tabular}}
\end{table}

\subsubsection{Excluded Properties and Scope}\label{excluded-properties-and-scope}

Coordinates, lengths, angle measures and layout are never facts. A redrawing may place the figure anywhere and at any size, and it is not asked to reproduce the printed proportions beyond what the relations themselves fix. Side and region relations other than containment, such as left\_of, above and same\_side, are computed on both the key and the drawing but are not scored; they serve only the contradiction check of Section 6.3.2, which detects a drawing that places a point on the wrong side of a line asserted elsewhere.

Colour-coded identification, as in Byrne's Euclid, lies outside the label vocabulary, so the keys of the 73 Byrne figures consist almost entirely of entities and relations.

The key is also a minimal set. A fact that follows from other facts by the closure rules, such as the endpoint memberships implied by an edge of a labelled polygon, is not stated again. A single drawing error therefore costs only one fact, and the chain of derived facts goes unpenalized.

\subsection{Dual-Writer Construction and Unanimity Rule}\label{dual-writer-construction-and-unanimity-rule}

To build the key without human annotation, we use two writer models from different providers, GPT-6.1 Sol (OpenAI) and Kimi K2.5 (Moonshot). A writer sees the crop, the words recognized on it and, for disambiguating labels only, the page text, so the key is the same for both tracks even though a Track A contestant must read the labels from the image alone. Each writer first lists the entities it sees and then the facts over those entities. After the two entity lists are matched by label, position and shared relations, with differently labelled objects never merged, each writer votes true, false or abstain on every fact in the combined list; a fact that a writer generated counts as supported by that writer unless the writer votes it false. Equation 1 states the rule that admits a fact to the key of figure \(i\):

\[K_i=\{\,f\in M_i:\ \mathrm{supp}_1(f)\wedge \mathrm{supp}_2(f)\wedge f\notin V_i\wedge \mathrm{args}(f)\cap A_i=\varnothing\wedge (f\in B_i\Rightarrow \mathrm{bind}(f))\,\} \tag{1}\]

\[\mathrm{supp}_w(f)\iff \bigl(f\in \mathrm{gen}_w\wedge v_w(f)\neq\mathrm{false}\bigr)\ \vee\ v_w(f)=\mathrm{true},\qquad v_w(f)\in\{\mathrm{true},\mathrm{false},\mathrm{abstain}\}\]

Here \(M_i\) is the closure-reduced union of both writers' facts for figure \(i\), and \(\mathrm{gen}_w\) is the set of facts that writer \(w\) generated. \(V_i\) is the set of facts that take part in a pattern of the exclusion table, such as a point that is both on and off a circle, so a fact is consistent when \(f\notin V_i\). \(A_i\) is the set of objects that the entity matching could not identify without doubt. \(B_i\) is the set of facts with an argument that must be located uniquely in the drawing, and \(\mathrm{bind}(f)\) holds when it is.

The rule has measured consequences. It keeps 84 percent of the entity facts, 79 percent of the label facts and 55 percent of the relation facts in the combined lists, and about 40 percent for tangent\_to, collinear and intersect\_at (Table S3). The key therefore over-represents facts that both writers see. Every contestant recovers relations that both writers generated more often than relations that only one writer generated, at 0.62--0.88 against 0.45--0.73 across models. A fact that one writer invents never enters the key, so a correct drawing is never marked wrong for omitting it. The screener and one writer come from the same provider (Table S2). Because unanimity makes the key incomplete by design, and a drawn fact absent from the key may still be true of the figure, we score recall of the key (Section 9.4 states the construct in full).

Two measurements made during development support the key. The first was made on the development key, which GPT-6 Sol wrote in place of GPT-6.1 Sol: on the 150 development figures, a combined fact is supported on average by 0.835 of the two writer families. The planned difference \(\delta^*\) of Section 6.5 was sized on the same key. The second was made earlier in development, when the writers were Claude Sonnet 5 and Kimi K2.5: on synthetic diagrams with known geometry, 170 of the 172 key facts that could be mapped to the ground truth were true; 76 percent of all key facts could be mapped, and the key held 51 percent of the true facts, which again shows a key that is accurate and incomplete.

\subsection{Item Gates and Key Statistics}\label{item-gates-and-key-statistics}

A carded figure is scored only when its key passes four gates:

\[|K^E_i|\ge 3,\qquad |K^R_i|\ge 3,\qquad |V_i|/|M_i|\le 0.10,\qquad |A_i|/|\mathcal E_i|\le 0.20 \tag{1a}\]

Here \(K^E_i\) and \(K^R_i\) are the entity and relation facts of the key, and \(\mathcal E_i\) is the set of matched objects. Together, the gates require enough entities and relations to score, few contradictions between the writers, and few objects that the matching could not identify. Of the 3,000 carded test figures, 2,507 pass, and Table S1 lists the exclusions by gate.

The keys contain 114,403 facts, a mean of 45.6 per figure. Because of the unanimity rule, the key retains \(\overline{|K_i|}/\overline{|M_i|} = 45.6/67.5\), or approximately two-thirds, of the combined facts.

\section{Scoring Protocol}\label{scoring-protocol}

\subsection{Compilation and Validity}\label{compilation-and-validity}

A fixed rule extracts the LaTeX block from the answer. A bare picture is wrapped in a fixed preamble, whereas a full document is compiled exactly as submitted, so a document class followed by a picture with no \texttt{document} environment is a compile error. We compile each answer once with LuaLaTeX.

Each answer receives one of 16 outcome codes (Table S4), and only the code OK yields a scored drawing. Equation 2 defines the validity indicator of model \(m\) on figure \(i\) under track \(t\):

\[\nu_i(m,t)=\begin{cases}1 & \text{if the outcome is OK}\\ 0 & \text{otherwise}\end{cases} \tag{2}\]

We define validity by the harness rule because a digitization workflow that compiles model output would apply the same rule.

\subsection{Vector Geometry Extraction and Alignment}\label{vector-geometry-extraction-and-alignment}

We read the compiled page as vector geometry from its drawing commands. The reader extracts path segments, circles and arcs fitted to runs of Bézier curves, filled regions and placed text, and types them by the conventions of the key (Section 5.1.1).

We then align the key to the drawing in four steps. Initial label matching binds a key object to the drawn object near a glyph bearing its label (the binding rule of Equation 3a); structural propagation extends the binding through shared relations; objects unique in their type are bound next; and the remainder is assigned by exact subgraph matching or optimal assignment. Where the alignment is ambiguous, a judge chooses among the candidate objects. A drawing with more than 60 primitives is not aligned geometrically, and all of its facts go to the judges. Because alignment uses relations as well as labels, a drawing with one misplaced letter keeps its relation credit and loses only that label fact.

\subsection{Fact Verification}\label{fact-verification}

\subsubsection{Geometric Predicate Tests}\label{geometric-predicate-tests}

Throughout, \(D\) is the diagonal of the bounding box of all visible paths and glyph boxes on the compiled page, and \(\varepsilon=0.01\). \(d(p,x)\) is the distance from a drawn point to a drawn object, \(\angle(\ell_1,\ell_2)\in[0^\circ,90^\circ]\) is the angle between two drawn linear objects, and \(o_c\) and \(r_c\) are the centre and radius of a circle. Equation 3 gives the core tests:

\[\mathrm{on}(p,x)\iff d(p,x)\le\varepsilon D \qquad(\text{likewise endpoint\_of, vertex\_of, center\_of}) \tag{3}\]

\[\mathrm{parallel}(\ell_1,\ell_2)\iff \angle(\ell_1,\ell_2)\le 1.5^\circ\ \wedge\ \mathrm{gap}(\ell_1,\ell_2)>\varepsilon D\]

\[\mathrm{perpendicular}(\ell_1,\ell_2)\iff |90^\circ-\angle(\ell_1,\ell_2)|\le1.5^\circ\]

\[\mathrm{tangent}(\ell,c,p)\iff \bigl|d(o_c,\ell)-r_c\bigr|\le 0.015\,r_c\ \wedge\ h(\ell,c)\le\varepsilon D\ \wedge\ \tau\in\mathrm{extent}(\ell)\ \wedge\ d(p,\tau)\le\varepsilon D\]

In the tangency test, \(h(\ell,c)=\sqrt{\max(0,\ r_c^2-d(o_c,\ell)^2)}\) is the half-chord, so a secant is not a tangent, and \(\tau\) is the touch point. In the parallelism test, the gap condition keeps overlapping collinear strokes from counting as parallel. Appendix B defines the other predicates. Labels are checked by a string rule:

\[\mathrm{label}(t,e)\iff \mathrm{norm}(t)=\mathrm{norm}(t')\ \wedge\ \mathrm{bind}(t')=e \tag{3a}\]

Here \(t'\) is a printed string in the drawing; \(\mathrm{bind}\) attaches \(t'\) to the nearest point entity within \(0.06\,D\), or else to the nearest other object within that radius, and \(\mathrm{norm}\) applies Unicode canonical composition (NFC), a map of confusable characters across Latin, Greek and Cyrillic, and Latin case folding. Case folding merges labels such as \(A\) and \(a\), so in the 149 keys that contain both, a label placed on the wrong one of the two objects can be credited.

Scaling or moving the page multiplies every distance, \(D\) and \(r_c\) by one factor and leaves angles unchanged, so every test is invariant to translation and uniform scale, as the prompt promises.

\subsubsection{Judge Panel and Verdict Rule}\label{judge-panel-and-verdict-rule}

A fact goes to the judges in three cases: when its predicate is judge-only (Table 3), when one of its objects is unbound, or when the drawing exceeds 60 primitives. The panel consists of Qwen3 VL 235B, Gemma 4 31B and Mistral Large 3, from families disjoint from the writers', and a judge never decides a contestant from its own family (Table S2). Judges see only the compiled drawing and answer every statement true or false under the same invariances as the task. For contestant \(m\) with eligible judge families \(J_m\) in fixed panel order, Equation 4 gives the verdict:

\[v(f)=\begin{cases}\text{agreement of the first two families; the third family decides a disagreement} & |J_m|=3\\ a_{j_1}(f),\ \text{the answer of the first eligible family} & |J_m|=2\end{cases} \tag{4}\]

Three contestants, Gemma 4 26B, Gemma 4 E2B and Ministral 3 14B, have two eligible families. For them, Qwen, the first eligible family, alone decides the judged facts; the second family's answer is recorded but does not decide. The two disagree on 18 to 27 percent of judged facts, and Section 8.5 bounds this rule by counting every such fact as absent and then as present.

Every fact receives one verdict from PRESENT, ABSENT, CONTRADICTED and EXEMPT. A fact is CONTRADICTED when the drawing asserts something the exclusion table makes incompatible with \(f\), for example when two lines the key makes parallel are drawn so that they cross, and EXEMPT when an object \(f\) needs is deterministically absent from the drawing (Section 6.4). To check judge stability, each judge re-answered its own development calls before and after scoring; agreement with its earlier answers was at least 0.867, against a required 0.8.

\subsection{Item and Benchmark Scores}\label{item-and-benchmark-scores}

For model \(m\) on track \(t\), the item score of figure \(i\) is the share of checkable key facts found PRESENT, multiplied by validity:

\[S_i(m,t)=\nu_i(m,t)\cdot\frac{|P_i|}{|K_i\setminus X_i|},\qquad P_i=\{f\in K_i\setminus X_i: v(f)=\mathrm{PRESENT}\} \tag{5}\]

\[X_i=\{f\in K_i:\ \exists\,x\in\mathrm{args}(f),\ x\ \text{absent}\}\ \cup\ \{f\in B_i:\ \text{no unique binding}\}\]

Because \(X_i\) depends on the drawing, a missing object costs its own entity fact once, while the relations and labels that mention it are set aside rather than charged again. On valid drawings, \(X_i\) removes 21 to 38 percent of relation facts across models; the fixed-denominator variant of Section 8.5 charges them instead by setting \(X_i=\varnothing\).

The benchmark score averages first within and then across source groups:

\[\bar S_k(m,t)=\frac1{n_k}\sum_{i\in k}S_i(m,t),\qquad H(m,t)=\frac1G\sum_{k=1}^{G}\bar S_k(m,t),\qquad G=87 \tag{6}\]

The target measure is the mean across source groups, the sampling units of the corpus (Section 3.3); validity rates, per-class recall and label-error shares are weighted by figure. For the 95 percent interval, the group means are treated as the observations (Ibragimov \& Müller, 2010):

\[H\pm t_{0.975,\,G-1}\,\frac{s_{\bar S}}{\sqrt G} \tag{7}\]

Here \(s_{\bar S}\) is the sample standard deviation of the group means, so the full set has 86 degrees of freedom and every subset uses its own \(G-1\).

Figure 3 traces the score through one figure and six drawings.

\begin{figure}[!htbp]
\centering
\includegraphics[width=\linewidth]{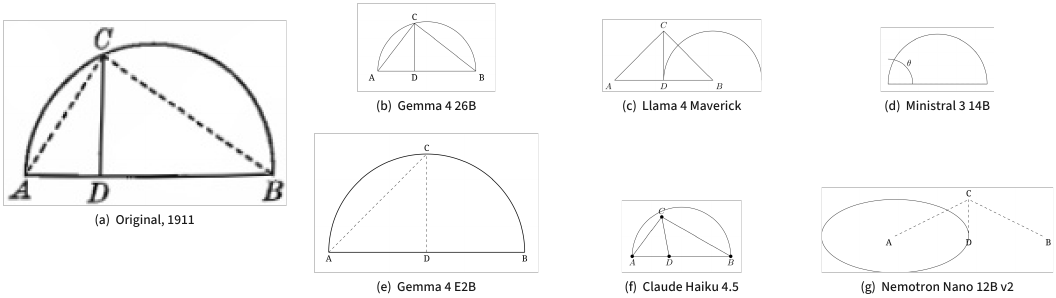}
\caption{Worked example (Track A; \emph{The Teaching of Geometry}, 1911; 40 key facts). (a) Original crop. (b)--(g) Compiled drawings of Gemma 4 26B, Llama 4 Maverick, Ministral 3 14B, Gemma 4 E2B, Claude Haiku 4.5 and Nemotron Nano 12B v2, at one common scale. S = found/checkable facts, with the number set aside in parentheses: (b) 0.93 = 25/27 (13), (c) 0.85 = 34/40 (0), (d) 0.07 = 1/14 (26), (e) 0.92 = 33/36 (4), (f) 0.85 = 34/40 (0), (g) 0.65 = 22/34 (6). Panel (b) draws the original's dashed lines solid and panel (e) keeps them; line style is not scored.}
\label{fig:3}
\end{figure}

\subsection{Statistical Comparison and Secondary Metrics}\label{statistical-comparison-and-secondary-metrics}

We compare two models by a paired t test on their group-mean differences:

\[\bar d_k=\bar S_k(m_1)-\bar S_k(m_2),\qquad T=\frac{\bar d}{s_d/\sqrt G},\qquad p=2\{1-F_{t,G-1}(|T|)\} \tag{8}\]

To correct for multiplicity over the 15 pairs of six models, we apply the Benjamini--Hochberg step-up procedure at \(q=0.10\) (Benjamini \& Hochberg, 1995), which rejects \(p_{(1)},\dots,p_{(k^*)}\) with \(k^*=\max\{k: p_{(k)}\le kq/15\}\). The rank band of model \(m\) is \([\,1+\#\text{models separated above } m,\ 6-\#\text{models separated below } m\,]\). Section 8.5 adds the Benjamini--Yekutieli adjustment, which is valid under any dependence among the tests (Benjamini \& Yekutieli, 2001).

We fixed the planned design difference \(\delta^*\), the minimum detectable difference of the planned design, on development data before the test run and pre-registered it both as the threshold for a practically meaningful difference between models and as the equivalence margin for page text:

\[\delta^*=(t_{0.975,29}+t_{0.80,29})\,\sigma_d\sqrt{\frac{1+(n_A-1)\rho}{N^*}}=0.140 \tag{9}\]

The planning values are \(\sigma_d=0.499\) and \(\rho=0.184\), estimated on the development figures, together with \(N^*=250\) figures in \(G^*=30\) groups, where \(n_A=\sum n_g^2/N\) gives \(n_A\approx8.3\). For page text, ``no meaningful pooled effect'' means

\[\mathrm{CI}_{95}\subset[-\delta^*,\delta^*], \tag{10}\]

which is the two one-sided tests at 0.025 each. We measure the stability of the model order by Kendall's \(\tau_b=(n_c-n_d)/\sqrt{(n_0-n_1)(n_0-n_2)}\) on point estimates, with \(n_0=15\) (Kendall, 1938), against a pre-registered threshold of 0.8. The leave-one-family-out score \(H^{(-j)}\) re-decides every judged fact from the remaining votes, with one vote deciding and a tie counting as not recovered. When the removed family is the contestant's own, no eligible judge is lost, and \(H^{(-j)}\) differs from the headline only in counting its two judges' disagreements as not recovered, the forced-absent bound.

The secondary metrics, defined in Appendix C, are Valid@1, the share of valid answers; \(H_{\text{valid}}\), Equation 6 over valid drawings only; the loss decomposition, which splits \(1-H\) between missing drawings and valid drawings; the common-valid difference, Equation 8 over figures both models draw validly; per-class and tier recall; Q5, the share of valid drawings that recover at least 80 percent of their entities yet contradict at least one relation; the contradicted share, the precision-side signal; and the judge share. Four variants modify the scoring rule. Geometry-confirmed \(H\) counts every judge-decided PRESENT as not recovered, a lower bound on \(H\) from geometry alone; fixed-denominator \(H\) sets \(X_i=\varnothing\); the forced bounds count every fact on which a two-family contestant's judges disagree as ABSENT and then as PRESENT; and the item-weighted mean is \(N^{-1}\sum_iS_i\).

\section{Experimental Setup}\label{experimental-setup}

The contestants are six low-cost vision-language models served by one inference provider, chosen to span five model families and a wide range of sizes: Gemma 4 26B and Gemma 4 E2B (Gemma Team, 2026), Llama 4 Maverick (Meta, 2025), Ministral 3 14B (Liu et al., 2026), Nemotron Nano 12B v2 VL (NVIDIA et al., 2025; Nemotron Nano 12B v2 below) and Claude Haiku 4.5 (Anthropic, 2025). Claude Haiku 4.5 is the one proprietary family. The inclusion rule also requires that every contestant have at least two eligible judge families, so that a second family's answer is available to bound the verdicts of the first (Section 6.3.2). We chose this cost class because a digitization workflow over the figures of whole libraries would run on models of this kind. The results describe these six models only.

Each model answers every figure once per condition, at temperature 0, under one shared limit on answer length and with identical prompt text and image rendering, which yields 30,084 answers over 2,507 figures, two tracks and six models. Tracks A and B of each model were scored in the same window. All Claude Haiku 4.5 answers and ten Gemma 4 26B answers were compiled by a later harness version, which differs only on a code path that had previously aborted, so no difference between the tracks of a model comes from the scoring rules.

The analysis plan, fixed before the test run, separates confirmatory from descriptive results. Three analyses are confirmatory: \(H\) on Track A with Benjamini--Hochberg bands; the pooled Track B − A difference under the equivalence rule of Equation 10; and \(\tau_b\ge0.8\) under judge-family removal. All other analyses are descriptive and are reported with group t intervals and without multiplicity correction unless stated. They comprise \(H_{\text{valid}}\) and the loss decomposition; the per-model and subset page-text differences, which report Benjamini--Hochberg-adjusted p values over their family of 18 tests; routing; relation tiers; the sensitivity variants of Section 8.5; and all strata.

\section{Results and Analysis}\label{results-and-analysis}

\subsection{Main Results and Model Ranking}\label{main-results-and-model-ranking}

On the image alone, the six models recover between 0.20 and 0.59 of the answer key (Table 4). Gemma 4 26B leads at 0.589 {[}0.529, 0.650{]}, ahead of Llama 4 Maverick at 0.524 {[}0.468, 0.581{]}. At \(q=0.10\) the Benjamini--Hochberg procedure separates 14 of the 15 pairs and yields five rank bands, leaving Gemma 4 E2B and Claude Haiku 4.5 tied (adjusted p 0.84). Eight of the separated pairs differ by more than \(\delta^*=0.14\), the planned design difference (Figure 4).

The separation of the top pair is narrow. Gemma 4 26B and Llama 4 Maverick differ by 0.065 and separate at \(q=0.10\) with an adjusted p of 0.089, but the separation disappears under the Benjamini--Yekutieli adjustment (adjusted p 0.295), when the facts on which the two judges disagree are counted as absent (adjusted p 0.22), or under restriction to geometry-confirmed facts (difference 0.006, adjusted p 0.84). It also fails without the Google judge family (adjusted p 0.16). That variant removes no judge from Gemma 4 26B, whose deciding family is Qwen; it instead counts the disagreements between Gemma 4 26B's two judges as not recovered. The top comparison also mixes two verdict rules, since Gemma 4 26B is decided by one judge family and Llama 4 Maverick by a panel of three. Table 6 reports the number of separated pairs under every variant.

\begin{table}[!htbp]
\centering
\caption{Main results (N = 2,507 figures, G = 87 source groups). A, B: Track A, Track B. H: group-mean item score (Equation 6). Band: Benjamini--Hochberg rank band on Track A at q = 0.10. Valid: share of answers that yield a drawing. \(H_{\text{valid}}\): H over valid drawings only. Invalid share: part of the Track A loss 1 − H due to missing drawings.}\label{tab:4}
{\footnotesize\setlength{\tabcolsep}{3pt}
\begin{tabular}{@{}llllllll@{}}
\toprule\noalign{}
Model & H, A & Band & H, B & Valid A (\%) & Valid B (\%) & \(H_{\text{valid}}\), A & Invalid share (\%) \\
\midrule\noalign{}
Gemma 4 26B & 0.589 & 1 & 0.511 & 72.7 & 68.9 & 0.811 & 69 \\
Llama 4 Maverick & 0.524 & 2 & 0.506 & 70.4 & 71.7 & 0.718 & 58 \\
Ministral 3 14B & 0.391 & 3 & 0.495 & 70.4 & 68.7 & 0.658 & 65 \\
Gemma 4 E2B & 0.309 & 4--5 & 0.325 & 60.2 & 57.8 & 0.582 & 67 \\
Claude Haiku 4.5 & 0.300 & 4--5 & 0.296 & 50.8 & 66.5 & 0.726 & 85 \\
Nemotron Nano 12B v2 & 0.197 & 6 & 0.266 & 51.0 & 47.2 & 0.548 & 78 \\
\bottomrule
\end{tabular}}
\end{table}

\begin{figure}[!htbp]
\centering
\includegraphics[width=\linewidth]{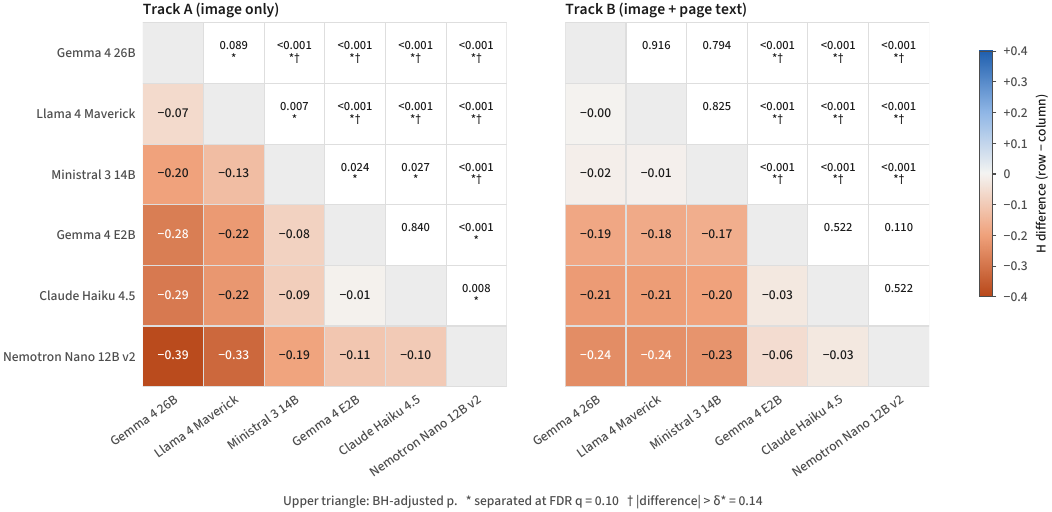}
\caption{Pairwise differences in H, row minus column, for Track A (left) and Track B (right). Upper triangle: Benjamini--Hochberg-adjusted p over the 15 pairs. *: separated at q = 0.10. †: absolute difference larger than δ* = 0.14, the planned design difference fixed on development data.}
\label{fig:4}
\end{figure}

\subsection{Compilation Validity and Loss Decomposition}\label{compilation-validity-and-loss-decomposition}

Most of each model's loss comes from answers that yield no drawing. On Track A these are 27 to 49 percent of answers and account for 58 to 85 percent of each model's loss; on Track B the share runs from 28 to 53 percent (Table 4; Figure 5).

The failure modes differ by model (Table S4). Nemotron is cut off at the answer-length limit on 651 of the 2,507 figures, and Llama 4 Maverick writes forbidden constructs on 292. Claude Haiku 4.5 fails to compile on 1,218, in 1,009 cases because it writes a document class and a picture with no \texttt{document} environment, whereas Ministral 3 14B and Gemma 4 E2B fail mostly through compile errors.

Drawings that do compile recover much more of the key, with \(H_{\text{valid}}\) ranging from 0.548 to 0.811, and on valid drawings alone the order differs from the headline. Over the 914 figures in 53 groups that both models draw validly, Claude Haiku 4.5 recovers 0.038 {[}0.001, 0.075{]} more than Llama 4 Maverick, so Haiku's fifth place comes from how often it produces a drawing. Vision2Code, which separates render success from reconstruction quality on modern material, reaches the same distinction (Periasami et al., 2026).

\begin{figure}[!htbp]
\centering
\includegraphics[width=\linewidth]{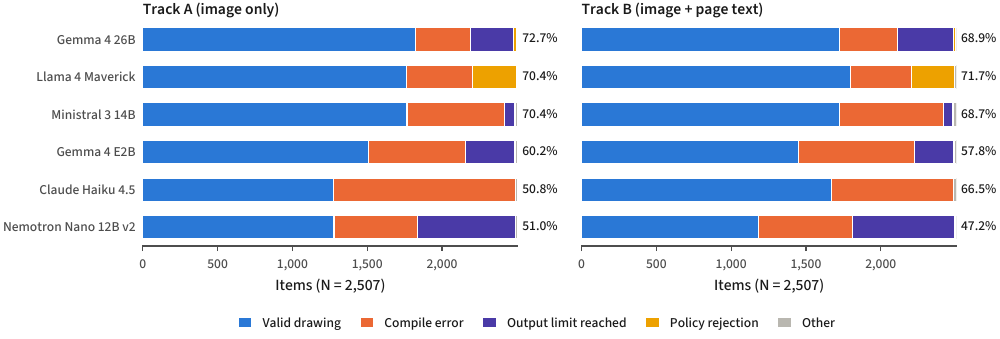}
\caption{Outcome of every answer by model and track (N = 2,507 figures per bar). Percentages give the share of valid drawings. Other: the remaining outcome codes of Table S4.}
\label{fig:5}
\end{figure}

\subsection{Recovery by Fact Class and Relation Tier}\label{recovery-by-fact-class-and-relation-tier}

Relations are the lowest class for every model, and within relations the depicted exact tier falls below the co-exact tier, again for every model (Figure 6). On Track A, entity recall ranges from 0.33 to 0.59, relation recall from 0.23 to 0.54 and label recall from 0.35 to 0.65. Labels are the highest class for five of the six models; for Claude Haiku 4.5, label recall (0.38) is level with entity recall (0.39). Under the fixed denominator, which charges the relations of missing objects, entities become the highest class for all six.

Co-exact relations are recovered at 0.24 to 0.55 and depicted exact relations at 0.17 to 0.44. Marked exact equalities occur on only 35 to 40 figures per model, so we draw no comparison from them.

Errors accumulate on relations because a relation requires both of its objects to bind and its geometry to satisfy the predicate test. Among valid drawings that recover at least 80 percent of their entities, 21 to 36 percent still contradict at least one relation (Q5). On the precision side, drawn facts that contradict the key make up 1.8 to 3.3 percent of checkable relations and 3.1 to 4.3 percent of checkable labels.

Label errors arise primarily from misattachment. On valid drawings the right string placed on the wrong object is the largest class of label error for every model, at 46 to 60 percent, ahead of wrong strings and missing labels.

Judges decide 33 to 41 percent of the verdicts on valid drawings. Most of these facts, 80 to 91 percent on Track A, involve an object that the alignment could not bind, and judges find 57 to 87 percent of the unbound entity facts present in the drawing.

\begin{figure}[!htbp]
\centering
\includegraphics[width=\linewidth]{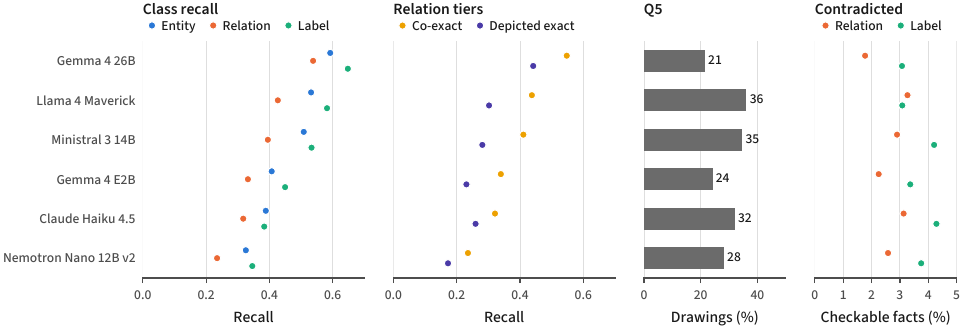}
\caption{What valid drawings recover on Track A. From left: recall by fact class; recall of co-exact and depicted exact relations; Q5, the share of valid drawings that recover at least 80 percent of their entities and still contradict at least one relation; and the contradicted share of checkable relation and label facts.}
\label{fig:6}
\end{figure}

\subsection{Effect of Page Text}\label{effect-of-page-text}

Pooled over the six models, page text leaves recovery unchanged within the pre-registered margin. The pooled difference B − A is +0.015, with a 95 percent interval from −0.010 to 0.040, which lies inside \(\pm\delta^*\) (Table 5; Figure 7) and also inside \(\pm0.04\).

Per model, over all figures and with group weighting, three differences remain statistically significant under Benjamini--Hochberg adjustment across the 18 per-model tests: Ministral 3 14B gains +0.104, Nemotron gains +0.069, and Gemma 4 26B decreases by −0.079 (intervals and p values in Table S5). The other three models change by less than 0.02. On Track B, the five bands of Track A merge into two (Figure 4).

Splitting the figures locates the source of these changes. On the 1,538 figures with page text, in 21 groups, the pooled difference is +0.009 {[}−0.009, 0.026{]}; there Claude Haiku 4.5 gains 0.160 {[}0.106, 0.214{]}, with 23 percentage points more valid drawings, and Nemotron changes by −0.076 {[}−0.126, −0.026{]}, with 9 points fewer. On the 969 text-free plates, in 70 groups, the image is identical across tracks and only the empty header is added, so, with one answer per model and condition, any change there combines the effect of the wrapper with run-to-run variation. Three models change, Ministral 3 14B by +0.134, Nemotron by +0.105 and Gemma 4 26B by −0.095. These shifts are concentrated in the 45 groups of three or fewer plates, where they reach +0.192, +0.158 and −0.148, against +0.030, +0.009 and 0.000 in the 25 larger groups (Table S5).

The per-model effects also depend on weighting, as item weighting gives Claude Haiku 4.5 a page-text effect of +0.120, against −0.004 under group weighting (Table S5). On figures valid in both conditions, pooled over figures with and without text, class recall changes by at most 0.055 for any model and class. Text therefore acts mainly on whether a drawing is produced, and the largest per-model shifts occur on small groups of plates to which no text is added.

\begin{table}[!htbp]
\centering
\caption{Page text: paired difference B − A in H (Track B minus Track A; G = 87, 21 and 70 groups); * marks a difference separated after Benjamini--Hochberg adjustment (q = 0.10), and pp is percentage points. Intervals and p values are in Table S5.}\label{tab:5}
{\footnotesize\setlength{\tabcolsep}{3pt}
\begin{tabular}{@{}lllll@{}}
\toprule\noalign{}
Model & All figures & With page text & Text-free plates & ΔValid, text (pp) \\
\midrule\noalign{}
Gemma 4 26B & −0.079* & −0.027 & −0.095* & −5.4 \\
Llama 4 Maverick & −0.018 & −0.007 & −0.026 & +3.1 \\
Ministral 3 14B & +0.104* & +0.023 & +0.134* & −5.9 \\
Gemma 4 E2B & +0.017 & −0.020 & +0.019 & −3.6 \\
Claude Haiku 4.5 & −0.004 & +0.160* & −0.048 & +23.1 \\
Nemotron Nano 12B v2 & +0.069* & −0.076* & +0.105* & −9.2 \\
Pooled & +0.015 & +0.009 & +0.015 & --- \\
\bottomrule
\end{tabular}}
\end{table}

\begin{figure}[!htbp]
\centering
\includegraphics[width=0.62\linewidth]{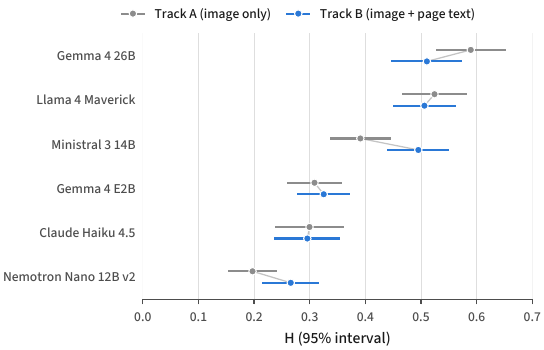}
\caption{H per model on Track A (image only) and Track B (image plus page text), with 95\% group-t intervals.}
\label{fig:7}
\end{figure}

\subsection{Ranking Stability and Stratified Analysis}\label{ranking-stability-and-stratified-analysis}

Across ten scoring and weighting variants, the point order of the six models is identical up to the order of Gemma 4 E2B and Claude Haiku 4.5, the pair the headline already ties (Table 6). The three tiers \{Gemma 4 26B, Llama 4 Maverick\}, \{Ministral 3 14B, Gemma 4 E2B, Claude Haiku 4.5\} and \{Nemotron Nano 12B v2\} are separated in every variant that tests separation.

Removing one judge family at a time leaves \(\tau_b=1.0\) for each family, which meets the confirmatory threshold of 0.8. Across the judging and scoring variants the number of bands ranges from three to six, and every \(\tau_b\) of 0.87 comes from the swap of Gemma 4 E2B and Claude Haiku 4.5 alone. Without the largest group, every score moves by at most 0.002; item weighting and the removal of the Numdam groups leave the point order unchanged; and the Benjamini--Yekutieli adjustment separates 13 of the 15 pairs.

The strata of Table 7 are descriptive. Geometry follows the headline order, but on graphs and plots (404 figures) Llama 4 Maverick scores 0.33 against 0.55 on geometry while Gemma 4 26B scores 0.53 against 0.60, so the gap between the top two models widens on this figure type. Ink touching the border changes \(H\) by less than 0.07 for every model. By period, five of the six models in this corpus score at least as high before 1850, on seven works, as in 1850--1930. The pre-1850 figures are mostly transcriber-prepared images, and that pathway scores highest for five of the six models, so the period comparison does not separate age from digitization.

\begin{table}[t]
\centering
\caption{Stability of the Track A comparison under scoring, weighting and multiplicity variants. $H$ per model; bands: distinct rank bands; separated: pairs of 15 separated at BH $q = 0.10$; top pair: adjusted $p$ for Gemma 4 26B against Llama 4 Maverick; $\tau_b$: Kendall's $\tau_b$ against the headline order. Item-weighted and without-Numdam rows report point order only; the Benjamini–Yekutieli row adjusts the headline $p$ values.}
\label{tab:stability}
\footnotesize\setlength{\tabcolsep}{2.8pt}
\begin{tabular}{@{}l rrrrrr cccc@{}}
\toprule
 & \multicolumn{6}{c}{$H$} & & & & \\ \cmidrule(lr){2-7}
Variant & \shortstack{Gemma 4\\26B} & \shortstack{Llama 4\\Maverick} & \shortstack{Ministral\\3 14B} & \shortstack{Gemma 4\\E2B} & \shortstack{Claude\\Haiku 4.5} & \shortstack{Nemotron\\Nano\\12B v2} & Bands & Sep. & \shortstack{Top pair\\$p_{\mathrm{adj}}$} & $\tau_b$ \\
\midrule
Headline & 0.589 & 0.524 & 0.391 & 0.309 & 0.300 & 0.197 & 5 & 14 & 0.089 & 1.00 \\
\midrule
Geometry-confirmed & 0.392 & 0.387 & 0.221 & 0.190 & 0.205 & 0.147 & 3 & 11 & 0.838 & 0.87 \\
Forced as absent & 0.571 & 0.524 & 0.355 & 0.298 & 0.299 & 0.197 & 3 & 11 & 0.220 & 0.87 \\
Forced as present & 0.607 & 0.524 & 0.406 & 0.360 & 0.300 & 0.197 & 6 & 13 & 0.035 & 1.00 \\
Fixed denominator & 0.553 & 0.487 & 0.355 & 0.263 & 0.276 & 0.165 & 5 & 14 & 0.081 & 0.87 \\
Without Google & 0.571 & 0.518 & 0.391 & 0.298 & 0.295 & 0.193 & 4 & 13 & 0.156 & 1.00 \\
Without Mistral & 0.589 & 0.500 & 0.355 & 0.309 & 0.282 & 0.185 & 6 & 13 & 0.022 & 1.00 \\
Without Qwen & 0.589 & 0.507 & 0.371 & 0.349 & 0.287 & 0.189 & 6 & 13 & 0.038 & 1.00 \\
Without largest group & 0.589 & 0.524 & 0.390 & 0.307 & 0.298 & 0.196 & 5 & 14 & 0.094 & 1.00 \\
Item-weighted & 0.571 & 0.489 & 0.460 & 0.379 & 0.361 & 0.295 & -- & -- & -- & -- \\
Without Numdam & 0.526 & 0.471 & 0.441 & 0.378 & 0.325 & 0.289 & -- & -- & -- & -- \\
Benjamini–Yekutieli & 0.589 & 0.524 & 0.391 & 0.309 & 0.300 & 0.197 & -- & 13 & 0.295 & -- \\
\bottomrule
\end{tabular}
\end{table}

\begin{table}[t]
\centering
\caption{Track A $H$ by stratum (descriptive). A stratum is scored only with at least 30 figures, 3 source groups and 150 key facts; --: below these minimums.}
\label{tab:strata}
\footnotesize\setlength{\tabcolsep}{4pt}
\begin{tabular}{@{}l rr rrrrrr@{}}
\toprule
 & & & \multicolumn{6}{c}{$H$} \\ \cmidrule(l){4-9}
Stratum & $n$ & $G$ & \shortstack{Gemma 4\\26B} & \shortstack{Llama 4\\Maverick} & \shortstack{Ministral\\3 14B} & \shortstack{Gemma 4\\E2B} & \shortstack{Claude\\Haiku 4.5} & \shortstack{Nemotron\\Nano\\12B v2} \\
\midrule
\multicolumn{9}{@{}l}{\textit{Figure type}} \\
\quad Geometry & 2,079 & 84 & 0.601 & 0.551 & 0.389 & 0.319 & 0.310 & 0.188 \\
\quad Graphs and plots & 404 & 34 & 0.531 & 0.331 & 0.373 & 0.256 & 0.227 & 0.186 \\
\addlinespace
\multicolumn{9}{@{}l}{\textit{Print condition}} \\
\quad Clear of the border & 1,683 & 76 & 0.603 & 0.535 & 0.387 & 0.340 & 0.314 & 0.222 \\
\quad Ink touches the border & 824 & 53 & 0.567 & 0.474 & 0.451 & 0.328 & 0.289 & 0.219 \\
\addlinespace
\multicolumn{9}{@{}l}{\textit{Period}} \\
\quad Before 1850 & 375 & 7 & 0.608 & 0.512 & 0.490 & 0.405 & 0.377 & 0.365 \\
\quad 1850–1930 & 1,965 & 82 & 0.591 & 0.525 & 0.387 & 0.305 & 0.302 & 0.190 \\
\quad After 1930 & 167 & 3 & 0.546 & 0.446 & 0.417 & 0.375 & 0.269 & 0.240 \\
\addlinespace
\multicolumn{9}{@{}l}{\textit{Repository}} \\
\quad Internet Archive & 948 & 16 & 0.542 & 0.448 & 0.459 & 0.357 & 0.319 & 0.291 \\
\quad Wikimedia Commons & 620 & 7 & 0.494 & 0.497 & 0.400 & 0.429 & 0.391 & 0.322 \\
\quad GDZ Göttingen & 294 & 8 & 0.451 & 0.479 & 0.439 & 0.351 & 0.291 & 0.257 \\
\quad Archival & 265 & 1 & -- & -- & -- & -- & -- & -- \\
\quad Project Gutenberg & 261 & 5 & 0.672 & 0.507 & 0.459 & 0.446 & 0.358 & 0.316 \\
\quad Numdam & 119 & 53 & 0.630 & 0.559 & 0.359 & 0.264 & 0.283 & 0.138 \\
\addlinespace
\multicolumn{9}{@{}l}{\textit{Digitization pathway}} \\
\quad Page-scan crop & 1,568 & 22 & 0.527 & 0.463 & 0.441 & 0.376 & 0.338 & 0.296 \\
\quad Repository plate image & 678 & 62 & 0.607 & 0.549 & 0.372 & 0.278 & 0.286 & 0.157 \\
\quad Transcriber-prepared image & 261 & 5 & 0.672 & 0.507 & 0.459 & 0.446 & 0.358 & 0.316 \\
\bottomrule
\end{tabular}
\end{table}

\section{Discussion}\label{discussion}

\subsection{Compilation Validity as the Primary Bottleneck}\label{compilation-validity-as-the-primary-bottleneck}

For a workflow that compiles model output, producing a valid drawing is the primary prerequisite. Missing drawings account for most of each model's loss, while valid drawings recover 0.55 to 0.81 of the key (Section 8.2). The failures are formal: answers cut off at the length limit, forbidden constructs, and an omitted \texttt{document} environment. Low executability is likewise the main failure of open models in chart-to-code generation (Zhao et al., 2025). Because the harness does not repair answers, these failures count against the model; a workflow that adds automated repair measures a different quantity, and the per-code counts of Table S4 show where repair would act. Claude Haiku 4.5 shows how much compilation validity matters. It ranks fifth on \(H\), yet on the figures that both it and Llama 4 Maverick draw validly it recovers more of the key.

\subsection{Depicted Exact Relations and Object Identity}\label{depicted-exact-relations-and-object-identity}

Parallels, perpendiculars, tangencies and collinearities are the relations every model recovers least, 0.06 to 0.14 below the co-exact tier. These are the relations that a scale drawing shows and that a proof text asserts (Section 2.1; Manders, 2008), and they are also the relations on which retrieval over a digitized collection depends, since a search for every figure in which a line touches a circle needs tangent\_to facts. Object identity poses a second challenge. A third of verdicts concern objects that the alignment could not bind, and judges find most such entity facts present in the drawing. Recovering identity (which drawn point represents A) is a distinct task, one that the label-attachment errors of Section 8.3 measure independently.

\subsection{Page Text and the Uninformative-Change Control}\label{page-text-and-the-uninformative-change-control}

Page text does not raise pooled recovery; its per-model effects run through validity and depend on how figures are weighted. In MathVerse, text lets models answer without reading the diagram (R. Zhang et al., 2024). In reconstruction the drawing must still be made; the text changes whether the drawing compiles and leaves alone what a valid drawing contains. We used the text-free plates as an uninformative-change control. For these plates the image and the key were identical and only an empty header was added, yet three models still shifted, mostly in small groups. With one answer per condition, the control cannot divide that shift between the wrapper and run-to-run variation, but it does set a floor, since a per-model effect of page text no larger than these shifts cannot be told apart from them. Reporting such a control with every added-input condition, with repeated answers where the budget allows, separates the effect of the input from the effects of the prompt and of sampling.

\subsection{Construct Validity and Transferability}\label{construct-validity-and-transferability}

\(H\) is the recall of the writer-agreed co-exact relations, depicted exact relations and printed labels of the printed witness, together with the entities on which they depend. It is decided on the model's own compiled geometry wherever the alignment binds the objects, and by judge models from other families otherwise; geometry-confirmed \(H\) reports the geometric part alone. Because \(H\) is recall, the contradicted share is its precision-side complement, and an application that indexes figures by their facts needs both. The construct is narrower than ``diagram understanding'' by design, in line with the case for benchmarks that state what they measure (Raji et al., 2021).

For RQ4, the order of the six models, up to the tied pair, does not depend on the judge families, the rule for judge disagreements, the denominator or the weighting. Geometry-confirmed \(H\), which ignores every judge-decided PRESENT, keeps the same order; the ranking therefore does not depend on judge verdicts of presence, although judges still choose bindings where the alignment is ambiguous. The separation of the top pair does depend on these choices, and Section 8.1 reports which.

Since the method requires neither source code nor human annotation, it can be run on printed collections whose figures contain labelled or co-exact structure, such as geometry in other languages and periods, technical plates, and the astronomical diagrams that Kalleli et al.~(2024) collect. The source books are in the public domain and plausibly in the pretraining data of the contestants (Sainz et al., 2023), and we did not test for contamination. The score checks the model's own drawing against facts, so a memorized image earns credit only when the model turns it into a program that satisfies them.

\subsection{Limitations}\label{limitations}

The study has three limitations. First, the answer keys and judge verdicts were not checked by human annotators on historical figures; their validity relies on the unanimity rule, on writer-family agreement of 0.835 on development figures, and on the synthetic check of Section 5.2, the latter two measured with earlier writer pairs. Second, the study evaluates six low-cost models and no frontier models, and the findings describe the six models of Section 7. Third, each model gave one answer per figure and condition, so run-to-run variation of the contestants was not measured; it enters every paired comparison, including the text-free control of Section 8.4, as unmodelled noise. Each gap has a direct extension: a stratified human audit of keys and verdicts, a frontier roster on the same protocol, and repeated answers on a subset of figures.

\section{Conclusion}\label{conclusion}

HistGramBench is a benchmark of 2,507 historical mathematical figures from 95 works printed between 1551 and 1933, paired with a protocol that checks a redrawn TikZ figure fact by fact on its compiled geometry and sends to judge models from other families only what geometry cannot bind. On six low-cost vision-language models, recovery of the answer key ranges from 0.20 to 0.59 on the image alone, and missing drawings account for most of each model's loss. Relations are the hardest facts, and depicted exact relations the hardest among them. Page text, pooled over models, leaves recovery unchanged. The point order of the models, up to one tied pair, and their three-tier structure persist across ten scoring and weighting variants, whereas the separation of the top two models depends on judging and multiplicity choices, and we report which. For the digitization of historical mathematics, the next measurable steps are reliable compilation, then exact relations and object identity, and the benchmark measures each of them. The figures, answer keys and scoring code are available so that further models can be evaluated under the same protocol and the same stability checks.

\section*{Data Availability}\label{data-availability}
\addcontentsline{toc}{section}{Data Availability}

HistGramBench v2 is released under the Apache License 2.0. The code, including the scoring harness and the scripts that compute every number in this paper, is available at \url{https://github.com/VietAlpha/histgrambench}. The dataset is available at \url{https://huggingface.co/datasets/VietAlphaLabs/HistGramBench}; access is granted on request until publication, after which the dataset will be open. Each figure is cited to its source work in the dataset's source catalogue.

\section*{Ethics Declaration}\label{ethics-declaration}
\addcontentsline{toc}{section}{Ethics Declaration}

The study involves no human subjects and no personal data. All source works are in the public domain; three entered on a recorded rights determination (Appendix D), and the two archival scans are held by the authors.

\section*{Funding}\label{funding}
\addcontentsline{toc}{section}{Funding}

This work was supported by Deerfield Academy and Amazon Web Services (AWS). The authors thank Deerfield Academy and AWS for their support.

\section*{References}\label{references}
\begingroup\everypar{\hangindent=1.5em\hangafter=1}
\addcontentsline{toc}{section}{References}

Anderson, P., Fernando, B., Johnson, M., \& Gould, S. (2016). SPICE: Semantic propositional image caption evaluation. In \emph{Computer Vision -- ECCV 2016} (pp.~382--398). Springer. \url{https://doi.org/10.1007/978-3-319-46454-1\_24}

Anthropic. (2025). \emph{Claude Haiku 4.5 system card}. \url{https://www-cdn.anthropic.com/7aad69bf12627d42234e01ee7c36305dc2f6a970.pdf}

Avigad, J., Dean, E., \& Mumma, J. (2009). A formal system for Euclid's \emph{Elements}. \emph{The Review of Symbolic Logic, 2}(4), 700--768. \url{https://doi.org/10.1017/S1755020309990098}

Baltacı, Z. S., Baena, R., Meng, F., Norindr, S., Somer, F., Husson, M., \& Aubry, M. (2026). Text region detection in historical astronomical diagrams. In \emph{Document Analysis and Recognition -- ICDAR 2026} (pp.~352--369). Springer. \url{https://doi.org/10.1007/978-3-032-36039-7\_21}

Belouadi, J., Ilg, E., Keuper, M., Tanaka, H., Utiyama, M., Dabre, R., Eger, S., \& Ponzetto, S. P. (2025). TikZero: Zero-shot text-guided graphics program synthesis. In \emph{Proceedings of the IEEE/CVF International Conference on Computer Vision (ICCV)} (pp.~17793--17806). \url{https://doi.org/10.1109/ICCV51701.2025.01653}

Belouadi, J., Lauscher, A., \& Eger, S. (2024). AutomaTikZ: Text-guided synthesis of scientific vector graphics with TikZ. In \emph{The Twelfth International Conference on Learning Representations (ICLR 2024)}. \url{https://proceedings.iclr.cc/paper\_files/paper/2024/hash/f7641940c7dd9e5de58c20e39586eb64-Abstract-Conference.html}

Belouadi, J., Ponzetto, S. P., \& Eger, S. (2024). DeTikZify: Synthesizing graphics programs for scientific figures and sketches with TikZ. In \emph{Advances in Neural Information Processing Systems 37} (pp.~85074--85108). \url{https://doi.org/10.52202/079017-2701}

Benjamini, Y., \& Hochberg, Y. (1995). Controlling the false discovery rate: A practical and powerful approach to multiple testing. \emph{Journal of the Royal Statistical Society: Series B (Methodological), 57}(1), 289--300. \url{https://doi.org/10.1111/j.2517-6161.1995.tb02031.x}

Benjamini, Y., \& Yekutieli, D. (2001). The control of the false discovery rate in multiple testing under dependency. \emph{The Annals of Statistics, 29}(4), 1165--1188. \url{https://doi.org/10.1214/aos/1013699998}

Blecher, L., Cucurull Preixens, G., Scialom, T., \& Stojnic, R. (2024). Nougat: Neural optical understanding for academic documents. In \emph{The Twelfth International Conference on Learning Representations (ICLR 2024)}. \url{https://proceedings.iclr.cc/paper\_files/paper/2024/hash/a39a9aceda771cded859ae7560530e09-Abstract-Conference.html}

Bo, W., Zhang, S., Sun, Y., Liu, J., Yao, Y., Du, J., He, W., Zou, K., Li, Z., \& Wang, J. (2026). \emph{Diagram-MMU: A multi-modal benchmark for scientific diagrams} {[}Preprint{]}. arXiv. \url{https://doi.org/10.48550/arXiv.2608.12262}

Bouche, T. (2008). Toward a digital mathematics library? In J. M. Borwein, E. M. Rocha, \& J. F. Rodrigues (Eds.), \emph{Communicating mathematics in the digital era} (pp.~47--73). A K Peters/CRC Press. \url{https://doi.org/10.1201/b10587-6}

Byrne, O. (1847). \emph{The first six books of the Elements of Euclid}. William Pickering. \url{https://doi.org/10.5479/sil.317414.39088000863027}

Cameron, A. C., \& Miller, D. L. (2015). A practitioner's guide to cluster-robust inference. \emph{Journal of Human Resources, 50}(2), 317--372. \url{https://doi.org/10.3368/jhr.50.2.317}

Committee on Planning a Global Library of the Mathematical Sciences. (2014). \emph{Developing a 21st century global library for mathematics research}. National Academies Press. \url{https://doi.org/10.17226/18619}

Cui, C., Sun, T., Liang, S., Gao, T., Zhang, Z., Liu, J., Wang, X., Zhou, C., Liu, H., Lin, M., Zhang, Y., Zhang, Y., Zheng, H., Zhang, J., Zhang, J., Liu, Y., Yu, D., \& Ma, Y. (2025). \emph{PaddleOCR-VL: Boosting multilingual document parsing via a 0.9B ultra-compact vision-language model} {[}Preprint{]}. arXiv. \url{https://doi.org/10.48550/arXiv.2510.14528}

Egiazarian, V., Voynov, O., Artemov, A., Volkhonskiy, D., Safin, A., Taktasheva, M., Zorin, D., \& Burnaev, E. (2020). Deep vectorization of technical drawings. In \emph{Computer Vision -- ECCV 2020} (pp.~582--598). Springer. \url{https://doi.org/10.1007/978-3-030-58601-0\_35}

Gebru, T., Morgenstern, J., Vecchione, B., Vaughan, J. W., Wallach, H., Daumé III, H., \& Crawford, K. (2021). Datasheets for datasets. \emph{Communications of the ACM, 64}(12), 86--92. \url{https://doi.org/10.1145/3458723}

Gemma Team. (2026). \emph{Gemma 4 technical report} {[}Preprint{]}. arXiv. \url{https://doi.org/10.48550/arXiv.2607.02770}

Hamami, Y., van der Kuil, M. N. A., Mumma, J., \& van der Ham, I. J. M. (2020). Cognitive processing of spatial relations in Euclidean diagrams. \emph{Acta Psychologica, 205}, Article 103019. \url{https://doi.org/10.1016/j.actpsy.2020.103019}

He, Q., Liu, X., Memon, H., Li, Z., Ma, Z., Cho, J., Ren, Z., Weld, D. S., \& Krishna, R. (2026). \emph{VFIG: Vectorizing complex figures in SVG with vision-language models} {[}Preprint{]}. arXiv. \url{https://doi.org/10.48550/arXiv.2603.24575}

Hilaire, X., \& Tombre, K. (2006). Robust and accurate vectorization of line drawings. \emph{IEEE Transactions on Pattern Analysis and Machine Intelligence, 28}(6), 890--904. \url{https://doi.org/10.1109/TPAMI.2006.127}

Hilbert, D. (1899). \emph{Grundlagen der Geometrie}. Teubner.

Ibragimov, R., \& Müller, U. K. (2010). \emph{t}-statistic based correlation and heterogeneity robust inference. \emph{Journal of Business \& Economic Statistics, 28}(4), 453--468. \url{https://doi.org/10.1198/jbes.2009.08046}

Kalleli, S., Trigg, S., Albouy, S., Husson, M., \& Aubry, M. (2024). Historical astronomical diagrams decomposition in geometric primitives. In \emph{Document Analysis and Recognition -- ICDAR 2024} (pp.~108--125). Springer. \url{https://doi.org/10.1007/978-3-031-70543-4\_7}

Kendall, M. G. (1938). A new measure of rank correlation. \emph{Biometrika, 30}(1--2), 81--93. \url{https://doi.org/10.1093/biomet/30.1-2.81}

Kim, J., Yang, R., \& Zhang, H. (2026). \emph{GeoBuildBench: A benchmark for interactive and executable geometry construction from natural language} {[}Preprint{]}. arXiv. \url{https://doi.org/10.48550/arXiv.2605.13167}

Kumar, V., Mishra, S., Hao, R., Malik, R., Broman, D., \& Demszky, D. (2025). \emph{DiagramIR: An automatic pipeline for educational math diagram evaluation} {[}Preprint{]}. arXiv. \url{https://doi.org/10.48550/arXiv.2511.08283}

Levchenko, M. (2025). Evaluating LLMs for historical document OCR: A methodological framework for digital humanities. In \emph{Proceedings of the First Workshop on Natural Language Processing and Language Models for Digital Humanities} (pp.~75--85). INCOMA Ltd.~\url{https://doi.org/10.26615/978-954-452-106-6-007}

Liu, A. H., Khandelwal, K., Subramanian, S., Jouault, V., Rastogi, A., Sadé, A., Jeffares, A., Jiang, A., Cahill, A., Gavaudan, A., Sablayrolles, A., Héliou, A., You, A., Ehrenberg, A., Lo, A., Eliseev, A., Calvi, A., Sooriyarachchi, A., Bout, B., . . . Ramzi, Z. (2026). \emph{Ministral 3} {[}Preprint{]}. arXiv. \url{https://doi.org/10.48550/arXiv.2601.08584}

Lu, P., Bansal, H., Xia, T., Liu, J., Li, C., Hajishirzi, H., Cheng, H., Chang, K.-W., Galley, M., \& Gao, J. (2024). MathVista: Evaluating mathematical reasoning of foundation models in visual contexts. In \emph{The Twelfth International Conference on Learning Representations (ICLR 2024)}. \url{https://proceedings.iclr.cc/paper\_files/paper/2024/hash/663bce02a0050c4a11f1eb8a7f1429d3-Abstract-Conference.html}

Lu, P., Gong, R., Jiang, S., Qiu, L., Huang, S., Liang, X., \& Zhu, S.-C. (2021). Inter-GPS: Interpretable geometry problem solving with formal language and symbolic reasoning. In \emph{Proceedings of the 59th Annual Meeting of the Association for Computational Linguistics and the 11th International Joint Conference on Natural Language Processing (Volume 1: Long Papers)} (pp.~6774--6786). \url{https://doi.org/10.18653/v1/2021.acl-long.528}

Manders, K. (2008). The Euclidean diagram (1995). In P. Mancosu (Ed.), \emph{The philosophy of mathematical practice} (pp.~80--133). Oxford University Press. \url{https://doi.org/10.1093/acprof:oso/9780199296453.003.0005}

Meta. (2025, April 5). \emph{The Llama 4 herd: The beginning of a new era of natively multimodal AI innovation} {[}Blog post{]}. \url{https://ai.meta.com/blog/llama-4-multimodal-intelligence/}

Miller, E. (2024). \emph{Adding error bars to evals: A statistical approach to language model evaluations} {[}Preprint{]}. arXiv. \url{https://doi.org/10.48550/arXiv.2411.00640}

Mumma, J. (2010). Proofs, pictures, and Euclid. \emph{Synthese, 175}(2), 255--287. \url{https://doi.org/10.1007/s11229-009-9509-9}

Netz, R. (1999). \emph{The shaping of deduction in Greek mathematics: A study in cognitive history}. Cambridge University Press. \url{https://doi.org/10.1017/CBO9780511543296}

Nikolaidou, K., Seuret, M., Mokayed, H., \& Liwicki, M. (2022). A survey of historical document image datasets. \emph{International Journal on Document Analysis and Recognition, 25}(4), 305--338. \url{https://doi.org/10.1007/s10032-022-00405-8}

NVIDIA, Deshmukh, A. S., Chumachenko, K., Rintamaki, T., Le, M., Poon, T., Mohseni Taheri, D., Karmanov, I., Liu, G., Seppanen, J., Chen, G., Sapra, K., Yu, Z., Renduchintala, A., Wang, C., Jin, P., Goel, A., Ranzinger, M., Voegtle, L., . . . Catanzaro, B. (2025). \emph{NVIDIA Nemotron Nano V2 VL} {[}Preprint{]}. arXiv. \url{https://doi.org/10.48550/arXiv.2511.03929}

Ouyang, L., Qu, Y., Zhou, H., Zhu, J., Zhang, R., Lin, Q., Wang, B., Zhao, Z., Jiang, M., Zhao, X., Shi, J., Wu, F., Chu, P., Liu, M., Li, Z., Xu, C., Zhang, B., Shi, B., Tu, Z., \& He, C. (2025). OmniDocBench: Benchmarking diverse PDF document parsing with comprehensive annotations. In \emph{Proceedings of the IEEE/CVF Conference on Computer Vision and Pattern Recognition (CVPR)} (pp.~24838--24848). \url{https://doi.org/10.1109/CVPR52734.2025.02313}

Panickssery, A., Bowman, S. R., \& Feng, S. (2024). LLM evaluators recognize and favor their own generations. In \emph{Advances in Neural Information Processing Systems 37} (pp.~68772--68802). \url{https://doi.org/10.52202/079017-2197}

Pasch, M. (1882). \emph{Vorlesungen über neuere Geometrie}. Teubner.

Peng, H. X., Kim, A., Li, A., Supasanya, C., Garg, S., \& Zhu, K. (2026). \emph{Solving is not drawing: A benchmark for diagrammatic reasoning in olympiad geometry} {[}Preprint{]}. arXiv. \url{https://doi.org/10.48550/arXiv.2608.18111}

Periasami, A. V., Wang, J., \& Dhingra, B. (2026). \emph{Vision2Code: A multi-domain benchmark for evaluating image-to-code generation} {[}Preprint{]}. arXiv. \url{https://doi.org/10.48550/arXiv.2605.11307}

Pfitzmann, B., Auer, C., Dolfi, M., Nassar, A. S., \& Staar, P. W. J. (2022). DocLayNet: A large human-annotated dataset for document-layout segmentation. In \emph{Proceedings of the 28th ACM SIGKDD Conference on Knowledge Discovery and Data Mining} (pp.~3743--3751). \url{https://doi.org/10.1145/3534678.3539043}

Raji, I. D., Bender, E. M., Paullada, A., Denton, E., \& Hanna, A. (2021). AI and the everything in the whole wide world benchmark. In \emph{Proceedings of the Neural Information Processing Systems Track on Datasets and Benchmarks} (Vol. 1). \url{https://datasets-benchmarks-proceedings.neurips.cc/paper\_files/paper/2021/hash/084b6fbb10729ed4da8c3d3f5a3ae7c9-Abstract-round2.html}

Rodriguez, J. A., Puri, A., Agarwal, S., Laradji, I. H., Rodriguez, P., Rajeswar, S., Vazquez, D., Pal, C., \& Pedersoli, M. (2025). StarVector: Generating scalable vector graphics code from images and text. In \emph{Proceedings of the IEEE/CVF Conference on Computer Vision and Pattern Recognition (CVPR)} (pp.~16175--16186). \url{https://doi.org/10.1109/CVPR52734.2025.01508}

Sainz, O., Campos, J. A., García-Ferrero, I., Etxaniz, J., Lopez de Lacalle, O., \& Agirre, E. (2023). NLP evaluation in trouble: On the need to measure LLM data contamination for each benchmark. In \emph{Findings of the Association for Computational Linguistics: EMNLP 2023} (pp.~10776--10787). \url{https://doi.org/10.18653/v1/2023.findings-emnlp.722}

Saito, K., \& Sidoli, N. (2012). Diagrams and arguments in ancient Greek mathematics: Lessons drawn from comparisons of the manuscript diagrams with those in modern critical editions. In K. Chemla (Ed.), \emph{The history of mathematical proof in ancient traditions} (pp.~135--162). Cambridge University Press.

Seo, M., Hajishirzi, H., Farhadi, A., Etzioni, O., \& Malcolm, C. (2015). Solving geometry problems: Combining text and diagram interpretation. In \emph{Proceedings of the 2015 Conference on Empirical Methods in Natural Language Processing} (pp.~1466--1476). \url{https://doi.org/10.18653/v1/D15-1171}

Si, C., Zhang, Y., Li, R., Yang, Z., Liu, R., \& Yang, D. (2025). Design2Code: Benchmarking multimodal code generation for automated front-end engineering. In \emph{Proceedings of the 2025 Conference of the Nations of the Americas Chapter of the Association for Computational Linguistics: Human Language Technologies (Volume 1: Long Papers)} (pp.~3956--3974). \url{https://doi.org/10.18653/v1/2025.naacl-long.199}

Suzuki, M., Tamari, F., Fukuda, R., Uchida, S., \& Kanahori, T. (2003). INFTY: An integrated OCR system for mathematical documents. In \emph{Proceedings of the 2003 ACM Symposium on Document Engineering} (pp.~95--104). \url{https://doi.org/10.1145/958220.958239}

Tantau, T. (2023). \emph{The TikZ and PGF packages: Manual for version 3.1.10}. CTAN. \url{https://ctan.org/pkg/pgf}

Verga, P., Hofstatter, S., Althammer, S., Su, Y., Piktus, A., Arkhangorodsky, A., Xu, M., White, N., \& Lewis, P. (2024). \emph{Replacing judges with juries: Evaluating LLM generations with a panel of diverse models} {[}Preprint{]}. arXiv. \url{https://doi.org/10.48550/arXiv.2404.18796}

Wu, C., Liang, Z., Ge, Y., Guo, Q., Lu, Z., Wang, J., Shan, Y., \& Luo, P. (2025). Plot2Code: A comprehensive benchmark for evaluating multi-modal large language models in code generation from scientific plots. In \emph{Findings of the Association for Computational Linguistics: NAACL 2025} (pp.~3006--3028). \url{https://doi.org/10.18653/v1/2025.findings-naacl.164}

Yang, C., Shi, C., Liu, Y., Shui, B., Wang, J., Jing, M., Xu, L., Zhu, X., Li, S., Zhang, Y., Liu, G., Nie, X., Cai, D., \& Yang, Y. (2025). ChartMimic: Evaluating LMM's cross-modal reasoning capability via chart-to-code generation. In \emph{The Thirteenth International Conference on Learning Representations (ICLR 2025)}. \url{https://openreview.net/forum?id=sGpCzsfd1K}

Zhang, L., Eger, S., Cheng, Y., Zhai, W., Belouadi, J., Moafian, F., \& Zhao, Z. (2025). ScImage: How good are multimodal large language models at scientific text-to-image generation? In \emph{The Thirteenth International Conference on Learning Representations (ICLR 2025)}. \url{https://proceedings.iclr.cc/paper\_files/paper/2025/hash/146b5b0feb14fe8e630669ad1faba25e-Abstract-Conference.html}

Zhang, M.-L., Yin, F., Hao, Y.-H., \& Liu, C.-L. (2022). Plane geometry diagram parsing. In \emph{Proceedings of the Thirty-First International Joint Conference on Artificial Intelligence (IJCAI-22)} (pp.~1636--1643). \url{https://doi.org/10.24963/ijcai.2022/228}

Zhang, R., Jiang, D., Zhang, Y., Lin, H., Guo, Z., Qiu, P., Zhou, A., Lu, P., Chang, K.-W., Qiao, Y., Gao, P., \& Li, H. (2024). MathVerse: Does your multi-modal LLM truly see the diagrams in visual math problems? In \emph{Computer Vision -- ECCV 2024} (pp.~169--186). Springer. \url{https://doi.org/10.1007/978-3-031-73242-3\_10}

Zhang, T., Wang, J., Yu, H., Xie, Q., Wang, J., \& Huang, H. (2026). MagicGeo: Training-free text-guided geometric diagram generation. \emph{Graphical Models, 145}, Article 101331. \url{https://doi.org/10.1016/j.gmod.2026.101331}

Zhang, X., Zhu, N., He, Y., Zou, J., Huang, Q., Jin, X., Guo, Y., Mao, C., Li, Y., Zhu, Z., Yue, D., Zhu, F., Wang, Y., Huang, Y., Wang, R., Qin, C., Zeng, Z., Xie, S., Luo, X., \& Leng, T. (2023). \emph{FormalGeo: An extensible formalized framework for olympiad geometric problem solving} {[}Preprint{]}. arXiv. \url{https://doi.org/10.48550/arXiv.2310.18021}

Zhao, X., Luo, X., Shi, Q., Chen, C., Wang, S., Liu, Z., \& Sun, M. (2025). ChartCoder: Advancing multimodal large language model for chart-to-code generation. In \emph{Proceedings of the 63rd Annual Meeting of the Association for Computational Linguistics (Volume 1: Long Papers)} (pp.~7333--7348). \url{https://doi.org/10.18653/v1/2025.acl-long.363}

Zheng, L., Chiang, W.-L., Sheng, Y., Zhuang, S., Wu, Z., Zhuang, Y., Lin, Z., Li, Z., Li, D., Xing, E. P., Zhang, H., Gonzalez, J. E., \& Stoica, I. (2023). Judging LLM-as-a-judge with MT-Bench and Chatbot Arena. In \emph{Advances in Neural Information Processing Systems 36} (pp.~46595--46623). \url{https://doi.org/10.52202/075280-2020}

Zou, B., Cai, M., Zhang, J., \& Lee, Y. J. (2024). VGBench: Evaluating large language models on vector graphics understanding and generation. In \emph{Proceedings of the 2024 Conference on Empirical Methods in Natural Language Processing} (pp.~3647--3659). \url{https://doi.org/10.18653/v1/2024.emnlp-main.213}

\par\endgroup
\section*{Appendix A. Prompt Templates}\label{appendix-a.-prompt-templates}
\addcontentsline{toc}{section}{Appendix A. Prompt Templates}

\textbf{Track A (contestant).} The full text, as sent with the figure crop:

\begin{quote}
You are given one image: a crop of a mathematical diagram from a historical printed book.

1. Task. Reconstruct the mathematical figure the original author intended; do not reproduce scan damage, paper texture, or print defects.

2. Output contract. Exactly one fenced block tagged ```latex, containing exactly one tikzpicture (optionally inside a standalone document). Any reasoning goes outside the block. Nothing after the block.

3. Allowlist. Packages: tikz, pgfplots, amsmath, amssymb, xcolor, tikz-3dplot. TikZ libraries: calc, intersections, angles, quotes, positioning, arrows.meta, decorations.markings, decorations.pathmorphing, patterns, through, shapes.geometric, shapes.misc, 3d, perspective, math, fit, backgrounds, babel. pgfplots libraries: fillbetween, polar, groupplots, colormaps, colorbrewer, statistics, ternary, smithchart, patchplots, units, dateplot, decorations.softclip. Lines before the tikzpicture may only load these packages and libraries, set pgfplots or TikZ options, define colours, or define macros. Explicitly not allowed: external, gnuplot-backed plotting, datavisualization, anything requiring shell escape, any font package.

4. Prohibitions. Shell escape, \textbackslash write18, \textbackslash directlua, \textbackslash input, \textbackslash include, \textbackslash openin, \textbackslash openout, filecontents, \textbackslash includegraphics, \textbackslash pdfximage, \textbackslash special, font-selection macros, external files, network access.

5. Not judged. Absolute coordinates, scale, line width, font, colour, small label offsets. The benchmark treats these as allowed variation: global translation; uniform scaling; line width; font family; colour; label offset within tolerance; canvas padding; DPI; source refactoring that yields the same render.

6. Script. Labels may be written literally in their original script (Chinese, Japanese, Vietnamese, Cyrillic, Greek, Latin); the compiler supplies covering fonts.
\end{quote}

\textbf{Track B (contestant).} The Track A text, followed by:

\begin{quote}
Context. After the image, the text below is surrounding text from the same page and may contain the figure's caption.
\end{quote}

and then up to 1,200 characters of page text, caption first. For a figure without page text, nothing follows the header.

\textbf{Writers (condensed).} Each writer receives the crop, the words recognized on it and the page text, with the instruction to use the page text only to disambiguate labels. It first lists the entities it sees, with labels where printed, using the 16 entity types of Section 5.1.1 and the conventions for points, straight strokes and polygons, and then lists facts over its own entities in the relation and label vocabulary. In a second pass, it votes true, false or abstain on every fact in the combined list of both writers.

\textbf{Judges.} Each judge sees only the compiled drawing and answers a list of statements:

\begin{quote}
You are checking a rendered mathematical diagram against a list of statements. You see ONE image: the rendered candidate diagram. You are not shown any source image, text or code. For every statement decide whether the rendered diagram asserts it (true) or not (false).
\end{quote}

The prompt then gives a convention note (the drawing conventions and the invariances of Section 4.2), the list of entities, in which a quoted name is a printed label and an identifier such as S1 is only a reference name, and the numbered statements. The judge answers with one JSON object holding exactly one true or false answer per statement, or an explicit refusal. When the alignment is ambiguous, a second prompt asks the judge to choose among candidate objects.

\section*{Appendix B. Fact Vocabulary}\label{appendix-b.-fact-vocabulary}
\addcontentsline{toc}{section}{Appendix B. Fact Vocabulary}

All 38 key predicates, with their argument types, tier and decision route. D is the diagonal of the bounding box of the compiled page and ε = 0.01. G: decided on the geometry when every argument binds, otherwise by the judges. J: always decided by the judges.

{\def\LTcaptype{none} 
\begin{longtable}[]{@{}
  >{\raggedright\arraybackslash}p{(\linewidth - 8\tabcolsep) * \real{0.2000}}
  >{\raggedright\arraybackslash}p{(\linewidth - 8\tabcolsep) * \real{0.2000}}
  >{\raggedright\arraybackslash}p{(\linewidth - 8\tabcolsep) * \real{0.2000}}
  >{\raggedright\arraybackslash}p{(\linewidth - 8\tabcolsep) * \real{0.2000}}
  >{\raggedright\arraybackslash}p{(\linewidth - 8\tabcolsep) * \real{0.2000}}@{}}
\toprule\noalign{}
\begin{minipage}[b]{\linewidth}\raggedright
Predicate
\end{minipage} & \begin{minipage}[b]{\linewidth}\raggedright
Arguments
\end{minipage} & \begin{minipage}[b]{\linewidth}\raggedright
Class or tier
\end{minipage} & \begin{minipage}[b]{\linewidth}\raggedright
Route
\end{minipage} & \begin{minipage}[b]{\linewidth}\raggedright
Test on the drawing
\end{minipage} \\
\midrule\noalign{}
\endhead
\bottomrule\noalign{}
\endlastfoot
point & --- & entity & G & a labelled or dotted point is bound \\
segment, line, ray & --- & entity & G & a straight stroke of the type is bound \\
circle, arc, ellipse & --- & entity & G & a fitted conic of the type is bound (Bézier fit, relative residual ≤ 0.01) \\
polygon & --- & entity & G & a chordless cycle of at most eight corners is bound \\
axis, arrow, angle marker, right-angle mark, length tick & --- & entity & G & an object of the type is bound by the alignment \\
region, curve, solid & --- & entity & J & --- \\
on & point, object & co-exact & G & d(p, x) ≤ εD \\
endpoint\_of & point, segment or arc & co-exact & G & d to the endpoint ≤ εD \\
vertex\_of & point, polygon or angle & co-exact & G & d to the vertex ≤ εD \\
center\_of & point, circle, arc or ellipse & co-exact & G & d to the fitted centre ≤ εD \\
edge\_of & segment, polygon & co-exact & G & both endpoints coincide with consecutive corners within εD \\
arc\_of & arc, circle & co-exact & G & arc on the circle; endpoints within εD \\
intersect\_at & object, object, point & co-exact & G & transversal crossing in the relative interior of both, directions differing by \textgreater{} 1.5°, within εD of the point \\
between & point, point, point & co-exact & G & middle point within εD of the outer segment and farther than εD from both ends \\
contains & closed object, closed object & co-exact & G & closed-in-closed over circle, ellipse, polygon, region \\
face\_of & polygon, solid & three-dimensional and other & J & --- \\
parallel & linear, linear & depicted exact & G & angle ≤ 1.5° and gap \textgreater{} εD \\
perpendicular & linear, linear & depicted exact & G & \textbar90° − angle\textbar{} ≤ 1.5° \\
tangent\_to & line or circle, circle or ellipse {[}, point{]} & depicted exact & G & \textbar d(o, ℓ) − r\textbar{} ≤ 0.015 r, half-chord ≤ εD, touch point within the line's extent (and within εD of the point) \\
collinear & ≥ 3 points & depicted exact & G & all within εD of one line \\
concyclic & ≥ 4 points & depicted exact & G & all within εD of one circle \\
equal\_length, equal\_angle & segments or angles & marked exact & J & counted only when the figure marks the equality \\
cross\_section\_of, hidden\_edge, asymptote\_of & various & three-dimensional and other & J & --- \\
label & string, object & label & G & normalized string equal; bound to the nearest point within 0.06 D, else the nearest other object \\
measure\_label & string, object & label & G & as label, for a printed measure \\
\end{longtable}
}

\textbf{Mapping to diagram-parsing annotations.} Inter-GPS literals PointLiesOnLine and PointLiesOnCircle map to on; Parallel and Perpendicular map to parallel and perpendicular. Inter-GPS literals that state numeric lengths and angle measures have no counterpart, because measures are never facts here. PGDP5K's geo2geo relations of incidence, parallelism and perpendicularity map to the same predicates, and its text2geo relation corresponds to label. HistGramBench adds betweenness, containment, tangency, collinearity and concyclicity as scored relations, labels as a scored class, typed arguments and closure reduction of the key.

\section*{Appendix C. Estimator Definitions}\label{appendix-c.-estimator-definitions}
\addcontentsline{toc}{section}{Appendix C. Estimator Definitions}

\textbf{Item score.} For model m, track t and figure i, Equation 5 with \(X_i\) as defined there. When \(\nu_i=0\), \(S_i=0\) and \(X_i=\varnothing\). Because \(|K^E_i|\ge3\) and entity facts are never in \(X_i\), the denominator is at least 3.

\textbf{Benchmark score and interval.} Equations 6 and 7, with \(G\) the number of source groups in the full set or subset. A subset needs \(G\ge6\) for an inferential claim. A subset whose group means have zero variance receives a point estimate and no interval.

\textbf{Paired comparison.} Equation 8. A pair with \(s_d=0\) or \(\bar d=0\) enters the Benjamini--Hochberg procedure with \(p=1\). Benjamini--Yekutieli multiplies the step-up thresholds by \(1/c(15)\) with \(c(15)=\sum_{k=1}^{15}1/k=3.318\).

\textbf{Valid@1.} \(N^{-1}\sum_i\nu_i(m,t)\), item-weighted.

\textbf{\(H_{\text{valid}}\).} Equation 6 over figures with \(\nu_i=1\), averaged over the groups that contain at least one such figure.

\textbf{Loss decomposition.} \(1-H=G^{-1}\sum_k\overline{(1-\nu)}_k+G^{-1}\sum_k\overline{\nu(1-S)}_k\); the invalid share of loss is the first term divided by \(1-H\).

\textbf{Common-valid difference.} Equation 8 restricted to figures with \(\nu_i(m_1)=\nu_i(m_2)=1\), over the groups that contain one.

\textbf{Per-class recall.} \(R^c=\frac{1}{|I_c|}\sum_{i\in I_c}\nu_i\,|P^c_i|/|K^c_i\setminus X^c_i|\), with \(I_c\) the figures having at least \(m_c\) checkable facts of class \(c\); \(m_E=3\), \(m_R=2\), \(m_L=1\).

\textbf{Tier recall.} As \(R^c\) with \(c\) a relation tier and \(m=1\).

\textbf{Q5.} Among valid drawings with entity recall at least 0.8, the share with at least one CONTRADICTED relation.

\textbf{Contradicted share.} CONTRADICTED facts over checkable facts on valid drawings, by class.

\textbf{Judge share and routing.} Judge-decided verdicts over all verdicts on valid drawings, including EXEMPT. Each judged fact is assigned to one of the three routing cases of Section 6.3.2.

\section*{Appendix D. Datasheet}\label{appendix-d.-datasheet}
\addcontentsline{toc}{section}{Appendix D. Datasheet}

Following Gebru et al.~(2021), this appendix summarizes the composition and provenance of the scored set; the full datasheet is released with the benchmark.

\textbf{Rights.} For every work, the release records the publication year with its evidence (catalogue record and title page), the source repository and URL, the checksum of the source file and its rights basis. Three scored works were printed after 1930 and entered on a recorded rights determination: the \emph{Shuxue cidian} (1933, Wikimedia Commons file SSID-11317445), Felix Klein's \emph{Vorlesungen über die hypergeometrische Funktion} (1933, GDZ PPN375394591) and Arthur Schoenflies and Max Dehn's \emph{Einführung in die analytische Geometrie der Ebene und des Raumes} (1931, GDZ PPN373602448).

\textbf{Archival works.} Two works enter from cleaned PDF scans held by the authors rather than from a public repository: George Bruce Halsted's \emph{The Elements of Geometry} (1885) and William Benjamin Smith's \emph{Introductory Modern Geometry of Point, Ray, and Circle} (1902). Figures were cut from the cleaned scans by the layout model, and the two works form one source group of 265 figures.

\clearpage\setcounter{table}{0}\setcounter{figure}{0}\renewcommand{\thetable}{S\arabic{table}}\renewcommand{\thefigure}{S\arabic{figure}}

\section*{Supplementary Material}\label{supplementary-material}
\addcontentsline{toc}{section}{Supplementary Material}

\begin{figure}[!htbp]
\centering
\includegraphics[width=0.62\linewidth]{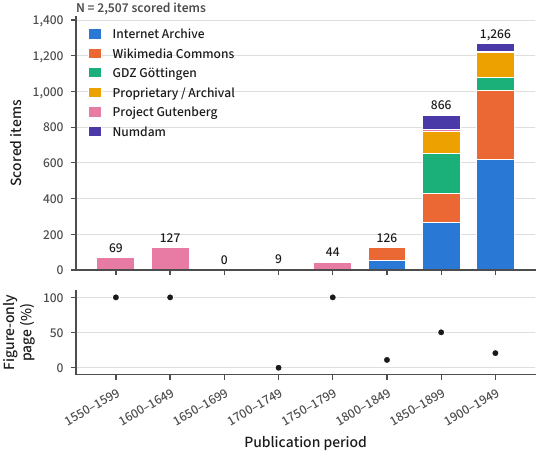}
\caption{Scored figures by period of publication and source repository.}
\label{fig:S1}
\end{figure}

\begin{table}[t]
\centering
\caption{Exclusion reasons with counts, over all candidate regions before the development/test split. A region split into panels is replaced by its panels.}
\label{tab:exclusions}
\small
\begin{tabular}{@{}l l r@{}}
\toprule
Reason & Code & Regions \\
\midrule
\multicolumn{3}{@{}l}{\textit{Image gates (3,256)}} \\
\quad Split into panels & \texttt{split\_into\_panels} & 703 \\
\quad Region too small & \texttt{box\_too\_small} & 563 \\
\quad Photographic & \texttt{photo\_grayscale} & 548 \\
\quad Extreme aspect ratio & \texttt{aspect\_extreme} & 287 \\
\quad Region too large & \texttt{box\_too\_large} & 287 \\
\quad Short strokes only & \texttt{no\_long\_strokes} & 217 \\
\quad Halftone & \texttt{halftone} & 204 \\
\quad Too dense & \texttt{ink\_too\_high} & 202 \\
\quad Text block & \texttt{text\_block} & 75 \\
\quad Physical size unknown & \texttt{physical\_size\_unknown} & 71 \\
\quad Prose in crop & \texttt{prose\_in\_crop} & 45 \\
\quad Solid block & \texttt{solid\_block} & 35 \\
\quad Too faint & \texttt{ink\_too\_low} & 19 \\
\addlinespace
\multicolumn{3}{@{}l}{\textit{Screening (8,609)}} \\
\quad Rejected by the screener & \texttt{verifier\_reject} & 8,609 \\
\addlinespace
\multicolumn{3}{@{}l}{\textit{Selection (2,043)}} \\
\quad Near-duplicate or second scan & \texttt{precard\_duplicate} & 1,530 \\
\quad Over the group's quota & \texttt{selection\_budget} & 513 \\
\addlinespace
\multicolumn{3}{@{}l}{\textit{Answer-key gates (556)}} \\
\quad Ambiguous entities & \texttt{card\_ambiguous\_entities} & 313 \\
\quad Too few key facts & \texttt{card\_insufficient\_facts} & 224 \\
\quad Writers contradict each other & \texttt{card\_contradiction\_rate} & 19 \\
\bottomrule
\end{tabular}
\end{table}

\begin{table}[t]
\centering
\caption{Model roles and families. A judge never decides a contestant from its own family; eligible judge families are listed in panel order.}
\label{tab:families}
\small
\begin{tabular}{@{}l l l l@{}}
\toprule
Role & Model & Family & Eligible judge families \\
\midrule
Writer & GPT-6.1 Sol & OpenAI &  \\
 & Kimi K2.5 & Moonshot &  \\
\addlinespace
Judge & Qwen3 VL 235B & Qwen &  \\
 & Gemma 4 31B & Google &  \\
 & Mistral Large 3 & Mistral &  \\
\addlinespace
Screener & GPT-6 Luna & OpenAI &  \\
\addlinespace
Contestant & Gemma 4 26B & Google & Qwen, Mistral \\
 & Llama 4 Maverick & Meta & Qwen, Google, Mistral \\
 & Ministral 3 14B & Mistral & Qwen, Google \\
 & Gemma 4 E2B & Google & Qwen, Mistral \\
 & Claude Haiku 4.5 & Anthropic & Qwen, Google, Mistral \\
 & Nemotron Nano 12B v2 & NVIDIA & Qwen, Google, Mistral \\
\bottomrule
\end{tabular}
\end{table}

\begin{table}[t]
\centering
\caption{Answer-key retention under the unanimity rule ($n =$ 114,403 retained facts). (a) Facts in the union of both writers and facts retained, by class; both: share of retained facts that both writers generated. (b) Share of checkable retained facts recovered on valid Track A drawings, by whether both writers or one generated the fact. (c) Retention by predicate.}
\label{tab:keyretention}
\small
\textbf{(a)}\\[2pt]
\begin{tabular}{@{}l rrrr@{}}
\toprule
Class & Union & Retained & Retention (\%) & Both (\%) \\
\midrule
Entity & 54,496 & 45,627 & 83.7 & 42.0 \\
Relation & 91,803 & 50,817 & 55.4 & 13.5 \\
Label & 22,806 & 17,959 & 78.7 & 69.9 \\
\bottomrule
\end{tabular}

\vspace{8pt}
\textbf{(b)}\\[2pt]
\begin{tabular}{@{}l rr rr rr@{}}
\toprule
 & \multicolumn{2}{c}{Entity} & \multicolumn{2}{c}{Relation} & \multicolumn{2}{c}{Label} \\
\cmidrule(lr){2-3} \cmidrule(lr){4-5} \cmidrule(l){6-7}
Model & Both & One & Both & One & Both & One \\
\midrule
Gemma 4 26B & 0.88 & 0.80 & 0.88 & 0.73 & 0.92 & 0.80 \\
Llama 4 Maverick & 0.81 & 0.74 & 0.76 & 0.58 & 0.85 & 0.69 \\
Ministral 3 14B & 0.75 & 0.75 & 0.68 & 0.55 & 0.81 & 0.72 \\
Gemma 4 E2B & 0.67 & 0.65 & 0.71 & 0.52 & 0.81 & 0.62 \\
Claude Haiku 4.5 & 0.82 & 0.76 & 0.79 & 0.60 & 0.85 & 0.70 \\
Nemotron Nano 12B v2 & 0.63 & 0.61 & 0.62 & 0.45 & 0.79 & 0.57 \\
\bottomrule
\end{tabular}

\vspace{8pt}
\textbf{(c)}\\[2pt]
\begin{tabular}{@{}l rr @{\hspace{2.5em}} l rr@{}}
\toprule
Predicate & Union & Kept (\%) & Predicate & Union & Kept (\%) \\
\midrule
\texttt{endpoint\_of} & 35,854 & 58 & \texttt{circle} & 1,194 & 72 \\
\texttt{label} & 21,949 & 79 & \texttt{center\_of} & 1,160 & 56 \\
\texttt{segment} & 21,385 & 84 & \texttt{arc} & 1,110 & 84 \\
\texttt{on} & 18,822 & 59 & \texttt{measure\_label} & 857 & 70 \\
\texttt{point} & 17,677 & 92 & \texttt{angle\_marker} & 582 & 84 \\
\texttt{edge\_of} & 9,200 & 64 & \texttt{region} & 436 & 88 \\
\texttt{intersect\_at} & 5,358 & 43 & \texttt{ray} & 420 & 55 \\
\texttt{parallel} & 5,128 & 48 & \texttt{length\_tick} & 344 & 72 \\
\texttt{line} & 3,844 & 80 & \texttt{arrow} & 312 & 71 \\
\texttt{polygon} & 2,912 & 68 & \texttt{asymptote\_of} & 305 & 1 \\
\texttt{perpendicular} & 2,909 & 53 & \texttt{ellipse} & 277 & 78 \\
\texttt{between} & 2,716 & 62 & \texttt{solid} & 273 & 58 \\
\texttt{vertex\_of} & 2,287 & 56 & \texttt{arc\_of} & 225 & 38 \\
\texttt{collinear} & 2,159 & 42 & \texttt{face\_of} & 209 & 4 \\
\texttt{tangent\_to} & 2,068 & 40 & \texttt{equal\_length} & 180 & 27 \\
\texttt{axis} & 1,942 & 87 & \texttt{right\_angle\_marked} & 122 & 2 \\
\texttt{curve} & 1,666 & 60 & \texttt{cross\_section\_of} & 82 & 10 \\
\texttt{hidden\_edge} & 1,559 & 22 & \texttt{concyclic} & 47 & 49 \\
\texttt{contains} & 1,491 & 44 & \texttt{equal\_angle} & 44 & 43 \\
\bottomrule
\end{tabular}
\end{table}

\begin{table}[t]
\centering
\caption{Outcome codes, with counts summed over the six models (15,042 answers per track). Only OK yields a scored drawing.}
\label{tab:outcomes}
\small
\begin{tabularx}{\linewidth}{@{}l l >{\raggedright\arraybackslash}X rr@{}}
\toprule
Stage & Code & Condition & Track A & Track B \\
\midrule
Extraction & \texttt{E\_NO\_OUTPUT} & Empty answer or refusal & 0 & 0 \\
 & \texttt{E\_TRUNCATED} & Answer cut off at the length limit & 1,332 & 1,373 \\
 & \texttt{E\_NO\_TIKZ} & No TikZ picture found & 14 & 12 \\
 & \texttt{E\_MULTI\_CANDIDATE} & More than one picture & 2 & 2 \\
Policy & \texttt{P\_POLICY\_REJECT} & Disallowed package, library or command & 325 & 329 \\
Compile & \texttt{C\_ERROR} & Compile error & 3,891 & 3,715 \\
 & \texttt{C\_TIMEOUT} & CPU time limit exceeded & 13 & 15 \\
 & \texttt{C\_WALLCLOCK} & Wall-clock kill & 0 & 0 \\
 & \texttt{C\_RESOURCE} & Memory, log, file-size or page cap & 5 & 5 \\
 & \texttt{C\_MISSING\_GLYPH} & Missing glyph & 24 & 23 \\
 & \texttt{C\_SANDBOX\_VIOLATION} & Sandbox violation & 1 & 0 \\
Output & \texttt{V\_MULTIPAGE} & More than one page & 0 & 1 \\
 & \texttt{V\_RASTER} & Raster image or embedded PostScript & 4 & 1 \\
 & \texttt{V\_EMPTY} & Blank drawing & 3 & 2 \\
 & \texttt{V\_OVERSIZE} & Page larger than the size cap & 15 & 17 \\
Valid & \texttt{OK} & Valid drawing & 9,413 & 9,547 \\
\bottomrule
\end{tabularx}
\end{table}

\begin{table}[t]
\centering
\caption{Page text: paired difference B $-$ A in $H$ by subset, with 95\% group-$t$ intervals ($df = G - 1$). Parentheses: BH-adjusted $p$ over the 18 per-model tests of the first three columns (separated at $q = 0.10$ when $\le 0.10$). Text-free plates are also split into groups of at most three and of more than three plates (estimates only). Item-wt.: item-weighted B $-$ A over all figures. $\Delta$Valid: change in the valid share, percentage points. 21 + 70 exceeds 87 because 4 groups contain both kinds of figure.}
\label{tab:pagetext}
\scriptsize\setlength{\tabcolsep}{3pt}
\begin{tabular}{@{}l ccccc c rr@{}}
\toprule
Model & \shortstack{All\\$G = 87$} & \shortstack{Page text\\$G = 21$} & \shortstack{Text-free\\$G = 70$} & \shortstack{Text-free, $\le$3\\$G = 45$} & \shortstack{Text-free, $>$3\\$G = 25$} & Item-wt. & \shortstack{$\Delta$Valid\\text} & \shortstack{$\Delta$Valid\\text-free} \\
\midrule
Gemma 4 & $-$0.079 (0.046) & $-$0.027 (0.391) & $-$0.095 (0.050) & $-$0.148 & +0.000 & $-$0.032 & $-$5.4 & $-$1.3 \\
26B & [$-$0.142, $-$0.015] & [$-$0.070, +0.016] & [$-$0.174, $-$0.016] &  &  &  &  &  \\\addlinespace[2pt]
Llama 4 & $-$0.018 (0.691) & $-$0.007 (0.764) & $-$0.026 (0.691) & $-$0.053 & +0.023 & +0.011 & +3.1 & $-$1.7 \\
Maverick & [$-$0.085, +0.048] & [$-$0.050, +0.035] & [$-$0.109, +0.058] &  &  &  &  &  \\\addlinespace[2pt]
Ministral & +0.104 (0.020) & +0.023 (0.691) & +0.134 (0.020) & +0.192 & +0.030 & +0.016 & $-$5.9 & +4.9 \\
3 14B & [+0.031, +0.177] & [$-$0.060, +0.107] & [+0.045, +0.224] &  &  &  &  &  \\\addlinespace[2pt]
Gemma 4 & +0.017 (0.691) & $-$0.020 (0.691) & +0.019 (0.691) & +0.034 & $-$0.007 & +0.001 & $-$3.6 & $-$0.5 \\
E2B & [$-$0.042, +0.075] & [$-$0.102, +0.062] & [$-$0.051, +0.090] &  &  &  &  &  \\\addlinespace[2pt]
Claude & $-$0.004 (0.906) & +0.160 (<0.001) & $-$0.048 (0.391) & $-$0.078 & +0.005 & +0.120 & +23.1 & +4.0 \\
Haiku 4.5 & [$-$0.070, +0.062] & [+0.106, +0.214] & [$-$0.125, +0.029] &  &  &  &  &  \\\addlinespace[2pt]
Nemotron & +0.069 (0.053) & $-$0.076 (0.020) & +0.105 (0.020) & +0.158 & +0.009 & $-$0.008 & $-$9.2 & +4.9 \\
Nano 12B v2 & [+0.010, +0.128] & [$-$0.126, $-$0.026] & [+0.035, +0.175] &  &  &  &  &  \\
\midrule
Pooled & +0.015 & +0.009 & +0.015 & -- & -- & -- & -- & -- \\
 & [$-$0.010, +0.040] & [$-$0.009, +0.026] & [$-$0.016, +0.046] &  &  &  &  &  \\
\bottomrule
\end{tabular}
\end{table}

\end{document}